\documentclass{article} % For LaTeX2e
\usepackage{iclr2022_conference,times}

\usepackage[utf8]{inputenc} % allow utf-8 input
\usepackage[T1]{fontenc}    % use 8-bit T1 fonts
\usepackage{hyperref}       % hyperlinks
\usepackage{url}            % simple URL typesetting
\usepackage{booktabs}       % professional-quality tables
\usepackage{amsfonts}       % blackboard math symbols
\usepackage{nicefrac}       % compact symbols for 1/2, etc.
\usepackage{microtype}      % microtypography
\usepackage{xcolor}         % colors

\usepackage{amsmath,amssymb}% math stuff
\usepackage{wrapfig}
\usepackage{enumitem}

\usepackage{subcaption}
\usepackage{listings}
\usepackage[normalem]{ulem}
\usepackage{csquotes}
\usepackage{bbm}            % For \mathbbm{1} for indicator functions
\usepackage{diagbox}
\usepackage{mathtools}      % For coloneqq (cute := sign)
\usepackage{xparse}         % For evalat
\usepackage{multirow}

\newcommand{\codeurl}{\url{https://github.com/pmichel31415/P-DRO}}

\usepackage[acronym,shortcuts,acronymlists={hidden}]{glossaries}
\newcommand{\eg}{\textit{e.g.}}
\newcommand{\ie}{\textit{i.e.}}
\newacronym{nlp}{NLP}{Natural Language Processing}
\newacronym{nmt}{NMT}{Neural Machine Translation}
\newacronym{smt}{SMT}{Statistical Machine Translation}
\newacronym{mt}{MT}{Machine Translation}
\newacronym{lm}{LM}{Language Modeling}
\newacronym{bpe}{BPE}{Byte-Pair Encoding}
\newacronym{dataset}{MTNT}{Machine Translation of Noisy Text}
\newacronym{mtnt}{MTNT}{Machine Translation of Noisy Text}
\newacronym{enfr}{\texttt{en-fr}}{English-French}
\newacronym{fren}{\texttt{fr-en}}{French-English}
\newacronym{enja}{\texttt{en-ja}}{English-Japanese}
\newacronym{jaen}{\texttt{ja-en}}{Japanese-English}
\newacronym{oov}{OOV}{out-of-vocabulary words}
\newacronym{rdb}{RDchrF}{relative decrease in chrF}
\newacronym{seq2seq}{seq2seq}{sequence-to-sequence}
\newacronym{nll}{NLL}{negative log likelihood}
\newacronym{dro}{DRO}{Distributionally robust optimization}
\newacronym{erm}{ERM}{empirical risk minimization}
\newacronym{mlp}{MLP}{multi-layer perceptron}
\newacronym{kl}{KL}{Kullback-Leibler}
\newacronym{p-dro}{P-DRO}{Parametric DRO}
\definecolor{Code}{rgb}{0,0,0}
\definecolor{Decorators}{rgb}{0.5,0.5,0.5}
\definecolor{Numbers}{rgb}{0.5,0,0}
\definecolor{MatchingBrackets}{rgb}{0.25,0.5,0.5}
\definecolor{Keywords}{rgb}{0,0,1}
\definecolor{self}{rgb}{0,0,0}
\definecolor{Strings}{rgb}{0,0.63,0}
\definecolor{Comments}{rgb}{0,0.63,1}
\definecolor{Backquotes}{rgb}{0,0,0}
\definecolor{Classname}{rgb}{0,0,0}
\definecolor{FunctionName}{rgb}{0,0,0}
\definecolor{Operators}{rgb}{0,0,0}
\definecolor{Background}{rgb}{0.98,0.98,0.98}
\lstdefinelanguage{Python}{numbers=left,numberstyle=\footnotesize,numbersep=1em,
xleftmargin=1em,framextopmargin=2em,framexbottommargin=2em,showspaces=false,
showtabs=false,showstringspaces=false,frame=l,
tabsize=4,
basicstyle=\ttfamily\small\setstretch{1},backgroundcolor=\color{Background},
commentstyle=\color{Comments}\slshape,
stringstyle=\color{Strings},morecomment=[s][\color{Strings}]{"""}{"""},morecomment=[s][\color{Strings}]{'''}{'''},morecomment=[l][\color{Strings}]{\#},
morekeywords={import,from,class,def,for,while,if,is,in,elif,else,not,and,or,print,break,continue,return,True,False,None,access,as,,del,except,exec,finally,global,import,lambda,pass,print,raise,try,assert},keywordstyle={\color{Keywords}\bfseries},
morekeywords={[2]@invariant,pylab,numpy,np,scipy},keywordstyle={[2]\color{Decorators}\slshape},emph={self},emphstyle={\color{self}\slshape},
}

\DeclareMathOperator*{\argmin}{arg\,min}

\newcommand{\KL}[2]{\text{KL}({#1} \Vert {#2})}
\DeclareMathOperator{\Ell}{\mathcal{L}}
\DeclareMathOperator{\E}{\mathbb{E}}
\NewDocumentCommand{\evalat}{sO{\big}mm}{%
  \IfBooleanTF{#1}
   {\mleft. #3 \mright|_{#4}}
   {#3#2|_{#4}}%
}

\newcommand{\founta}{\textsc{FDCL18}}
\title{Distributionally Robust Models \\ with Parametric Likelihood Ratios}

\author{Paul Michel \\
Centre Sciences des Donn\'ees \\
\'{E}cole normale sup\'{e}rieure PSL\\
Paris, 75005, France \\
{\footnotesize\texttt{pmichel31415@gmail.com}} \\
\And
Tatsunori Hashimoto \\
Computer Science Department \\
Stanford University \\
Stanford, CA 94305, USA \\
{\footnotesize\texttt{thashim@stanford.edu}} \\
\And
Graham Neubig \\
School of Computer Science \\
Carnegie Mellon University \\
Pittsburgh, PA 15213, USA \\
{\footnotesize\texttt{gneubig@cs.cmu.edu}}
}

\iclrfinalcopy % Uncomment for camera-ready version, but NOT for submission.
\begin{document}

\maketitle

\begin{abstract}
  As machine learning models are deployed ever more broadly, it becomes increasingly important that they are not only able to perform well on their training distribution, but also yield accurate predictions when confronted with distribution shift. The Distributionally Robust Optimization (DRO) framework proposes to address this issue by training models to minimize their expected risk under a collection of distributions, to imitate test-time shifts. This is most commonly achieved by instance-level re-weighting of the training objective to emulate the likelihood ratio with possible test distributions, which allows for estimating their empirical risk via importance sampling (assuming that they are subpopulations of the training distribution). However, re-weighting schemes in the literature are usually limited due to the difficulty of keeping the optimization problem tractable and the complexity of enforcing normalization constraints. In this paper, we show that three simple ideas -- mini-batch level normalization, a KL penalty and simultaneous gradient updates -- allow us to train models with DRO using a broader class of parametric likelihood ratios. In a series of experiments on both image and text classification benchmarks, we find that models trained with the resulting parametric adversaries are consistently more robust to subpopulation shifts when compared to other DRO approaches, and that the method performs reliably well with little hyper-parameter tuning.\footnote{Code to reproduce our experiments can be found at \codeurl{}}
  
\end{abstract}

\section{Introduction}

It is well acknowledged that modern neural network based machine learning models tend to underperform when they are evaluated on data distributions that differ from the one they were trained on. For example, machine learning model performance has been observed to degrade under train-test mismatch in topics \citep{gururangan-etal-2020-dont}, demographics \citep{blodgett2016demographic,amodei2016deep,hovy2015tagging,grother2019face}, geographic regions \citep{koh2020wilds}, and even data collection processes \citep{beery2018recognition,zech2018variable,michel18mtnt}. In particular, these models often perform poorly when evaluated on subpopulations, domains that are present but underrepresented in their training data \citep{sagawa2019distributionally}, and they can latch on to spurious correlations \citep{mccoy2019right}. This has problematic real-world consequences: when such models are deployed at large, this representation disparity can, for example, unfairly affect minority groups \citep{buolamwini2018gender,hashimoto2018fairness}.

This behaviour can largely be attributed to the \ac{erm} principle which underlies the majority of learning algorithms used in practice. In \ac{erm}, models are trained to minimize the average loss over a finite sample from a fixed training distribution \citep{vapnik1992principles}, as a proxy for the expected loss on a random example drawn from the fixed, albeit unknown, data distribution. This favors models which perform well \emph{on average} on a fixed training set, as opposed to models which would perform equally well on a variety of subpopulations that better reflects the diverse set of distributions that can be encountered at test time.
On the other hand, \ac{dro} proposes an appealing alternative to \ac{erm}. In \ac{dro}, models are trained to minimize their worst case risk (or an empirical estimate thereof computed on a finite sample, via \eg{} importance weighting) under a pre-determined family of distributions $\mathcal{Q}$, called the ``uncertainty set'' (or ``ambiguity set''):
\begin{equation}\label{eq:dro_loss}
    \mathcal L_{\text {DRO}}(\theta)=\max_{q\in\mathcal Q}\E_{(x,y)\sim q}\ell_\theta(x,y).
\end{equation}
In the absence of explicit information about the subpopulations of interest (which would naturally define $\mathcal Q$), it is up to the practitioner to carefully define this uncertainty set. This has been the subject of much work in the literature (see \citet{rahimian2019distributionally} for a survey). Recently, \citet{michel2021modeling} proposed P-DRO, a promising approach where the uncertainty set is defined by a parametric family of generative models, which allows for more flexibility in defining the uncertainty set. P-DRO shows significant improvement over comparable baselines, but it suffers from several drawbacks. First, it presupposes the availability of generative models capable of outputting exact densities, which limits its field of application to modalities where such models are readily available (such as language models in NLP). Second, it is challenging to use in practice due to its reliance on a number of hyper-parameters and approximations to the objective function.

In this paper, we propose a new approach for \ac{dro}, called RP-DRO, based on a key modification of the P-DRO algorithm: instead of modeling the worst-case distributions directly, we parametrize the likelihood ratio between the training distribution and the worst-case distribution. This removes the dependency on an unwieldy generative model, making the method useful for more applications. 
While likelihood ratio formulations of DRO have been tried in prior work \citep{sagawa2019distributionally}, we show that they are particularly effective for parametric, neural network based adversaries. Our approach relies on three simple ideas: a mini-batch level normalization strategy to enforce likelihood ratio constraints, a penalty-form of the KL divergence uncertainty set, and the use of simultaneous gradient updates for training. RP-DRO consistently achieves equal or better robust subpopulation accuracy compared to P-DRO and other baselines on a variety of standard benchmarks in image and text classification. In addition, we find it is both faster than P-DRO and depends on fewer hyper-parameters. Additional ablation experiments demonstrate that both our mini-batch normalization strategy and simultaneous gradient updates are necessary for high performance. Finally, we perform experimental analyses to shed light on the advantages brought by parametric adversaries compared to their nonparametric counterparts.

\vspace{-0.5em}
\section{Background}

In the following, we consider a model parametrized by $\theta\in \mathbb R^{d_\text{model}}$. Our goal is to find a model which minimizes the loss function $\ell_\theta(x, y)$ on pairs of inputs and outputs $(x,y)\in \mathcal{X}\times\mathcal{Y}$. For instance, $x$ might represent images and $y$ a categorical label. Parameters $\theta$ are estimated on a training dataset $\mathcal{D}_\text{train}=\{(x_i,y_i)\}_{i=1\ldots N_{\text{train}}}$ which we assume to be drawn from a training distribution $p$.

The DRO optimization problem with uncertainty set $\mathcal{Q}$ is
\begin{equation}\label{eq:dro_problem}
    \min_{\theta}\max_{q\in\mathcal Q}\E_{q}\ell_\theta(x, y).
\end{equation}
Note that the DRO loss in Eq. \ref{eq:dro_loss} is the inner maximum of the DRO problem, and it provides an upper bound on the expected loss of the model under any distribution in the uncertainty set $\mathcal Q$. This motivates the use the minimizer of the min-max game in Eq. \ref{eq:dro_problem} as a robust model. We refer to the solution of the inner maximum as the “adversary” from now on

However this objective is only useful insofar that (1) $\mathcal Q$ covers test distributions of interest (corresponding to different domains, demographics, etc.) and (2) $\mathcal Q$ is not overly pessimistic. To fulfil this second condition, there should exist some model $\theta^*$ that achieves low loss simultaneously on the test distribution as well as $\mathcal Q$. This often requires that $\mathcal Q$ only contain distributions that are covariate shifts of the test distribution, \ie{} that are such that the conditional distribution $q(y\mid x)$ coincides with that of training distribution $p(y\mid x)$.

\vspace{-0.5em}
\subsection{nonparametric DRO}

There is substantial existing work on \emph{nonparametric} formulations of DRO, where $\mathcal{Q}$ is expressed as a divergence ball centered at the training distribution. This includes $f$-divergences \citep{ben2013robust,hu2013kullback,faury2020distributionally}, Wasserstein/IPM \citep{sinha2017certifying,husain2020distributional}, moment constraints \citep{delage2010distributionally,nguyen2020robust}, and CVaR \citep{fan2017learning,curi2020adaptive,levy2020large} based uncertainty sets. These nonparametric approaches are appealing as they require very little domain-specific knowledge, have well-understood theory \citep{duchi2018learning}, and optimization procedures (\eg{} \citet{hu2013kullback} for KL constraints and \citet{levy2020large} for $\chi^2$ and CVaR constraints).

Unfortunately, nonparametric DRO algorithms suffer from being overly pessimistic. Their uncertainty sets tend to include distributions that are exceedingly difficult to learn, or not representative of real-world distribution shifts. Furthermore, they often cannot enforce even basic constraints such as covariate shift structures \citep{DuHa20dis, hu2018does}. Group-structured DRO uncertainty sets \citep{sagawa2019distributionally} overcome some of these challenges, but require significant domain expertise to pre-specify target subpopulations that a model should be robust to.

\subsection{Parametric DRO}

Parametric DRO \citep{michel2021modeling} is a method for DRO in which the uncertainty set $\mathcal Q$ is defined as a family of parametric generative models, which avoids the extreme pessimism of nonparametric DRO without the explicit specification of subpopulations. Specifically, given a generative model $q_\psi$ parameterized by $\psi\in\mathbb R^{d_{\text{adv}}}$, the KL-constrained parametric DRO objective can be written as follows:
\begin{equation}\label{eq:pdro}
    \min_{\theta}\max_{\substack{\psi\\\KL{q_\psi}{p}\leq \kappa}} \mathbb E_{(x,y)\sim q_\psi}\ell(x,y,\theta).
\end{equation}
As demonstrated by \citet{michel2021modeling}, P-DRO yields significant improvements over its  nonparametric counterpart. However, the difficulty of optimizing Eq. \ref{eq:pdro} directly results in a number of approximations and additional hyper-parameters that are hard to tune.

In addition, a central drawback of P-DRO is that it necessitates training an auxiliary generative model of the data. This can be difficult for several reasons. First, this limits the applicability of the method to domains with generative models that allow for exact probability computations. Moreover, even when such generative models are available, they are often more computationally demanding than their discriminative counterparts. In language models for instance, probabilities for sequences of text are obtained by iteratively producing conditional probabilities over all tokens in the vocabulary. This additional step results in considerable computational overhead compared to discriminative models.

\section{Parametric Likelihood Ratio}
\label{sec:method}

\vspace{-0.5em}
\subsection{DRO as a Likelihood Ratio Optimization Problem}
\label{sec:likelihood_ratio}
In the situation that all distributions in $\mathcal Q$ are absolutely continuous with respect to $p$ (i.e. for all measurable subset $A\subset\mathcal X\times \mathcal Y$, all $q\in \mathcal Q$, $q(A)>0$ only if $p(A)>0$) the inner maximum in Eq. \ref{eq:dro_problem} can be rewritten purely as a function of the likelihood ratio $\frac{q}{p}$
\begin{equation}
\E_{(x,y)\sim q}\ell_\theta(x, y)=\E_{(x,y)\sim p}\frac{q(x,y)}{p(x,y)}\ell_\theta(x,y).
\end{equation}
Such absolute continuity assumptions are standard in $f$-divergence and group DRO methods, which both rely upon re-weighting the training distributions. In fact, the KL divergence constraint in P-DRO presupposes absolute continuity.

This suggests that the inner maximum can be re-written as an optimization problem on functions $r: \mathcal{X}\times\mathcal{Y}\longrightarrow\mathbb R_+$ within the uncertainty set $\mathcal R\in \{r\mid pr\in \mathcal Q\}$
\begin{equation}\label{eq:ratio_dro_problem}
    \min_{\theta}\max_{r\in\mathcal R}\E_{(x,y)\sim p}r(x,y)\ell_\theta(x, y).
\end{equation}
This reparametrization of the problem will allow us to replace a parametric family of generative models with a parametric family over probability ratios.

\subsection{Ratio-based P-DRO}

The likelihood ratio formulation described above is appealing for P-DRO because it enables the use of discriminative style neural architectures for parametrizing the ratio $r$, which opens up many more options for defining the parametric uncertainty set. Specifically, we can set the adversary to be any parametric function $r_\psi : \mathcal X\times Y\longrightarrow \mathbb R^+$ verifying $\E_{x,y\sim p}r_\psi(x, y)=1$. The key insight that we use to realize our proposed method is that we do not need to restrict the choice of adversaries to those that implicitly satisfy this normalization condition (\ie{} generative models). Instead, we can pick any adversary and treat normalization as an additional constraint (the ``normalization constraint'').

Note that in this case, the KL constraint takes the simple form $\KL{pr_\psi}{p}=\E_{pr_\psi}\log\frac{pr_\psi}{p}=\E_{p}r_\psi\log r_\psi$. The final min-max problem, which we dub ratio-based P-DRO (RP-DRO), is:
\begin{equation}
    \min_{\theta}\max_{\substack{\psi\\\E_{p}r_\psi\log r_\psi\leq \kappa\\\mathbb E_{p} r_\psi=1}} \underbrace{\mathbb E_{(x,y)\sim {p}}r_\psi(x,y)\ell_\theta(x,y)}_{\mathcal L_{\text{RP-DRO}}}.
\end{equation}
As in P-DRO, we can look for equilibria of this differentiable min-max game by performing simultaneous gradient updates \citep{singh2000nash} to $\theta$ and $\psi$ in directions $-\nabla_\theta \mathcal{L}_{\text{RP-DRO}}$ and $+\nabla_\psi \mathcal{L}_{\text{RP-DRO}}$ respectively. Although finding global equilibria is not guaranteed in high dimensional non-convex settings \citep{balduzzi2018mechanics}, empirical evidence suggests that models trained in this manner still reach useful solutions \citep{michel2021modeling}.

In experiments, we adopt an exponential parametrization $r_\psi(x,y)\propto e^{f_\psi(x,y)}$ where $f_\psi$ is the output of any parametric model with values in $\mathbb R$. Similarly to P-DRO, we do not explicitly enforce the KL constraint (due to the difficulty of projecting onto the KL ball), and instead we relax it in the form of a term $\tau\E_{p}r_\psi\log r_\psi$ added to the loss function. The regularization strength $\tau$ is treated as a hyper-parameter.

\vspace{-0.5em}
\subsection{Enforcing the normalization constraint}
\label{sec:normalization}

In addition to the KL constraint, RP-DRO necessitates that $r_\psi$ satisfies a normalization constraint $\E_{p} r_\psi=1$ to ensure that ${p}r_\psi$ is a proper probability distribution over $\mathcal D_\text{train}$. If that were not the case, the adversary $r_\psi$ could artificially increase the weighted expectation $\E_{p}r_\psi\ell_\theta$ by assigning a total weight greater than 1 to the entire dataset.

Existing methods for ratio based DRO such as \cite{sagawa2019distributionally} achieve this by either projecting $r_\psi$ onto the set $\{r\mid \E_pr=1\}$ after each gradient update on $\psi$, or by directly parametrizing the ratio as $r_\psi(x, y)=e^{f_\psi(x, y)}/\E_pe^{f_\psi}$. Unfortunately, these solutions are computationally infeasible in practical scenarios with large datasets. Indeed, they necessitate computing the entire expectation over $p$, which can be costly when each $f_\psi(x, y)$ is the output of a neural model.

We propose two simple, yet effective solutions for addressing this issue in the context of mini-batch training where we can only compute $f_\psi$ for small number of samples ${(x_1, y_1),\ldots,(x_n, y_n)}$ at each step of training.

\paragraph{Self-normalization} is inspired by the idea of ``self-normalization'' developed for globally normalized structured prediction models \citep{andreas2015accuracy,goyal2019empirical}. It consists in adding a relaxation of the normalization constraint to the objective. Specifically, following \citet{goyal2019empirical} we add a squared penalty on the log normalizer at the mini-batch level. Ignoring the KL penalty, this regularized objective takes the following empirical form:
\begin{equation}
    {\hat{\Ell}_{\text{self-norm}}}(\theta, \psi)=\frac 1 n\sum_{i=1}^nr_\psi(x_i, y_i)\ell_\theta(x_i, y_i) - \beta\log\left(\frac 1 n\sum_{i=1}^nr_\psi(x_i, y_i)\right)^2.
\end{equation}
The hyper-parameter $\beta$ controls the regularization strength. Intuitively, this penalizes adversaries that assign too much (or too little) total weight to the mini-batch. However, the choice of an optimal $\beta$ adds an additional degree of freedom to RP-DRO, which suggests our second option as a simpler alternative.

\paragraph{Batch-level normalization} consists of using the normalized ratio at the mini-batch level by setting
\begin{equation}
    \tilde r_\psi(x_i,y_i) = \frac{e^{f_\psi(x_i, y_i)}}{\sum_{j=1}^ne^{f_\psi(x_j, y_j)}}
\end{equation}
for each sample $(x_i, y_i)$ in the mini-batch. An obvious downside of this approach is that the weight of each sample now depends on the mini-batch it was sampled from. This can be problematic for small batch sizes: as an extreme example, for a batch size of 1, this normalization scheme assigns the same weight of 1 to every sample, making the objective equivalent to ERM.

However, mini-batch approximations have proven effective for other forms of DRO \citep{hu2018does,levy2020large} and there is some evidence that they can yield accurate estimates for higher batch sizes \citep{cortes2010learning}. In practice we find that this approach yields good results for commonly used batch sizes, is generally more stable than the self-normalization penalty, and does not introduce the additional hyper-parameter $\beta$. In most of our experiments, we adopt this approach unless specified otherwise. In that case, the empirical RP-DRO objective on a mini-batch becomes
\begin{equation}
    \hat{\Ell}_{\text{batch-level norm}}(\theta, \psi)=\underbrace{\sum_{i=1}^n\tilde{r}_\psi(x_i, y_i)\ell_\theta(x_i, y_i)}_{\text{expected loss under }pr_\psi} - \tau \underbrace{\sum_{i=1}^n \tilde{r}_\psi(x_i, y_i)\log \tilde{r}_\psi(x_i, y_i)}_{\text{KL penalty}}.
\end{equation}
The KL term serves to penalize ratios that deviate too much from~$1$ (Note that the penalty is subtracted because we are maximizing with respect to $\psi$). The only hyper-parameter that needs to be tuned is the KL regularization strength $\tau$.

\vspace{-0.5em}
\section{Experiments}
\label{sec:experiments}

\vspace{-0.5em}
\subsection{Datasets}
\label{sec:datasets}

We perform experiments on four datasets: two text classification datasets used in \citet{michel2021modeling}, BiasedSST and \founta{}, and two image classification datasets from \citet{sagawa2019distributionally}, Waterbirds and CelebA. Specific details for each dataset follow these previous works, as described below:

\paragraph{BiasedSST} is based on the SST-2 sentiment classification dataset \citep{radford2018improving}, but modified to introduce spurious correlation between a distractor token (``So,'') and positive labels in around 95\% of the dataset. In this setting models trained with ERM can very easily learn this spurious correlation, which hurts performance on the small subpopulation that does not suffer from this bias.
\paragraph{\founta{}} A toxicity detection dataset of tweets labeled as  \textit{hateful} ($5\%$), \textit{abusive} ($27\%$), \textit{normal} ($54\%$) and \textit{spam} ($14\%$).
The group-DRO problem is formulated by partitioning the evaluation data along labels as dialectal annotation obtained automatically with an off-the shelf classifier \citep{blodgett2016demographic,sap2019risk}. In particular these dialects align closely with self-reported race, and \citet{sap2019risk} found that machine learning models trained on such toxicity detection datasets tend to exhibit bias towards certain labels depending on the demographics of the tweets' authors, particularly with minorities. In order to report more reliable accuracy numbers, all groups containing less than 100 samples are aggregated when computing test accuracies.

\paragraph{Waterbirds} An image classification dataset where the task is to predict ``land bird'' or ``water bird'' with the confounding factor of the background; most water (resp. land) bird pictures have water (resp. land) on the background.

\paragraph{CelebA} A popular face recognition dataset originally published by \citet{liu2015faceattributes}. The group-DRO problem is formulated as a task of predicting the hair color (``Blond'' or ``Dark'') across groups formed by the combination of the label and the (binary) gender of the subject. Due to the spurious correlation between blond hair/female and dark hair/male, models trained with ERM tend to achieve lower accuracies on underrepresented groups such as ``blond-haired male'' which totals only 0.85\% of the training data.

\subsection{Experimental Setting}
\label{sec:expt_setting}

On BiasedSST and \founta{} we follow \citet{michel2021modeling} and train a BiLSTM and BERT-base model \cite{devlin2018bert} respectively. On the image classification datasets we train Resnet-50 architectures \citep{he2016deep} pre-trained on ImageNet \citep{deng2009imagenet} as in \citet{sagawa2019distributionally}.

\begin{table*}[t]

\caption{\label{tab:results_text} Robust and average test accuracies on the Biased SST and \founta{} datasets.  Underlined numbers indicates statistically significant difference compared to \ac{erm} ($p$-value$<0.05$). Bold numbers indicates the best number in each column (barring Oracle DRO).}
\centering
\begin{tabular}{l|cc|cc}
\toprule
  & \multicolumn{2}{c}{Biased SST} &\multicolumn{2}{c}{\founta{}}\\
 & Robust & Average & Robust & Average\\
\midrule

ERM & { 2.15}  {\scriptsize$\pm$ 0.97} & { \bf 95.09}  {\scriptsize$\pm$ 0.16} & { 19.57}  {\scriptsize$\pm$ 7.00} & {\bf  81.56}  {\scriptsize$\pm$ 0.26}\\
Topic DRO & \underline{ 5.18}  {\scriptsize$\pm$ 1.46} & { 95.00}  {\scriptsize$\pm$ 0.10} & { 16.48}  {\scriptsize$\pm$ 5.46} & \underline{ 80.49}  {\scriptsize$\pm$ 0.49} \\
NonParam-$\chi^2$& \underline{ 4.65}  {\scriptsize$\pm$ 1.61} & { 95.10}  {\scriptsize$\pm$ 0.22} & { 15.70}  {\scriptsize$\pm$ 2.56} & { 81.55}  {\scriptsize$\pm$ 0.23} \\
NonParam-CVaR & \underline{ 24.42}  {\scriptsize$\pm$ 6.82} & \underline{ 89.03}  {\scriptsize$\pm$ 3.73} & { 11.19}  {\scriptsize$\pm$ 7.64} & \underline{ 69.19}  {\scriptsize$\pm$ 8.28} \\
NonParam-KL & \underline{ 28.11}  {\scriptsize$\pm$ 2.16} & \underline{ 92.45}  {\scriptsize$\pm$ 1.55} & \underline{ 17.54}  {\scriptsize$\pm$ 6.41} & \underline{ 81.20}  {\scriptsize$\pm$ 0.11} \\
P-DRO & \underline{ 34.98}  {\scriptsize$\pm$ 9.39} & \underline{ 84.21}  {\scriptsize$\pm$ 2.11} & {30.25}  {\scriptsize$\pm$ 10.13} & { 79.91}  {\scriptsize$\pm$ 1.41} \\
RP-DRO & \underline{\bf 50.70}  {\scriptsize$\pm$ 7.33} & \underline{ 86.60}  {\scriptsize$\pm$ 2.96} & \underline{ \bf 53.52}  {\scriptsize$\pm$ 1.66} & \underline{ 76.62}  {\scriptsize$\pm$ 1.43} \\\midrule
Oracle DRO & \underline{ 67.71}  {\scriptsize$\pm$ 3.03} & \underline{ 77.91}  {\scriptsize$\pm$ 4.49} & \underline{ 55.23}  {\scriptsize$\pm$ 3.97} & \underline{ 72.43}  {\scriptsize$\pm$ 2.61} \\
\bottomrule
\end{tabular}
\end{table*}

\begin{table*}[t]

\caption{\label{tab:results_images} Robust and average test accuracies on the Waterbirds and CelebA datasets. Underlined numbers indicates statistically significant difference compared to \ac{erm} ($p$-value$<0.05$). Bold numbers indicates the best number in each column (barring Oracle DRO).}
\centering
\begin{tabular}{l|cc|cc}
\toprule
  & \multicolumn{2}{c}{Waterbirds} &\multicolumn{2}{c}{CelebA}\\
 & Robust & Average & Robust & Average\\
\midrule

ERM & { 68.32}  {\scriptsize$\pm$ 2.02} & { 89.23}  {\scriptsize$\pm$ 0.36} & { 40.33}  {\scriptsize$\pm$ 2.29} & { \bf 95.89}  {\scriptsize$\pm$ 0.05} \\
NonParam-$\chi^2$ & { 68.54}  {\scriptsize$\pm$ 0.65} & { 89.56}  {\scriptsize$\pm$ 0.61} & { 41.78}  {\scriptsize$\pm$ 2.03} & { 95.87}  {\scriptsize$\pm$ 0.07} \\
NonParam-CVaR & \underline{ 44.71}  {\scriptsize$\pm$ 14.27} & \underline{ 71.94}  {\scriptsize$\pm$ 10.30} & { 36.00}  {\scriptsize$\pm$ 7.50} & \underline{ 94.63}  {\scriptsize$\pm$ 0.58} \\
NonParam-KL & \underline{ 72.21}  {\scriptsize$\pm$ 0.95} & \underline{ \bf 90.54}  {\scriptsize$\pm$ 0.72} & { 43.33}  {\scriptsize$\pm$ 3.58} & \underline{ 95.72}  {\scriptsize$\pm$ 0.10} \\
RP-DRO & \underline{ \bf 73.49}  {\scriptsize$\pm$ 3.01} & { 90.15}  {\scriptsize$\pm$ 0.74} & \underline{ \bf 55.78}  {\scriptsize$\pm$ 9.15} & { 93.10}  {\scriptsize$\pm$ 3.87} \\\midrule
Oracle DRO & \underline{ 85.60}  {\scriptsize$\pm$ 0.95} & { 89.12}  {\scriptsize$\pm$ 1.20} & \underline{ 89.22}  {\scriptsize$\pm$ 0.90} & \underline{ 92.59}  {\scriptsize$\pm$ 0.40} \\
\bottomrule
\end{tabular}
\vspace{-1em}
\end{table*}

Since the adversary in RP-DRO can be any discriminative architecture, we opt for the natural solution of using a similar architecture for this model. For instance on BiasedSST, we take $f_\psi(x,y)$ as the raw logit output by a BiLSTM architecture identical to that of the classifier (without the final softmax layer). We do the same for the other datasets, with the exception of \founta{} where we use a smaller DistillBERT model \citep{sanh2019distilbert} for efficiency. We use the same learning rate and optimizer for both model and adversary and only vary the KL penalty weight $\tau\in\{0.001,0.01,0.1,1.0\}$. We use a batch size of 64 for BiasedSST and \founta{} and 32 for Waterbirds and CelebA. We perform optimal stopping using the Minmax criterion proposed in \citet{michel2021modeling}: every epoch $T$, we determine the best model by explicitly solving a greedy approximation of the min-max game between all $T$ previously checkpointed adversaries and models on the validation dataset $\mathcal D_{\text{valid}}$.
\begin{equation}\label{eq:valid_minmax}
    \theta^*=\argmin_{\theta_{i=1\ldots T}}\max_{\psi_{j=1\ldots T}} \frac 1 {|D_{\text{valid}}|} \sum_{x,y\in D_{\text{valid}}}\tilde{r}_{\psi_j}(x, y)\ell_{\theta_i}(x,y)
\end{equation}
A similar strategy is applied for hyper-parameter selection. Importantly, we substitute the 0-1 loss for $\ell_\theta$ in Eq. \ref{eq:valid_minmax} (only for validation) as we found in preliminary experiments on BiasedSST that it consistently produced better results.

We compare our results to 5 different methods experimented with by \citet{michel2021modeling}:
\begin{itemize}[nolistsep,leftmargin=*]
    \item \textbf{ERM}: standard training to minimize the average loss.
    \item \textbf{NonParam}: nonparametric DRO with various uncertainty sets. We report results for uncertainty set constrained by KL divergence \citep{hu2013kullback,hu2018does}, $\chi^2$ divergence and CVaR \citep{levy2020large}, respectively referred to as \textbf{NonParam-KL}, \textbf{NonParam-$\chi^2$} and \textbf{NonParam-CVaR}. In all of these variants, the inner maximization problem has an analytical solution where $q(x,y)$ depends on some function of $\ell_\theta(x,y)$. Consequently, examples with high loss are up-weighted.
    \item \textbf{Topic-DRO}: a variation on Topic CVaR \citep{oren2019distributionally} using the online algorithm from \citet{sagawa2019distributionally} to minimize the worst case loss on a collection of pseudo domains automatically generated via topic modeling.\footnote{This baseline was inaccurately referred to as ``Topic CVaR'' in \citet{michel2021modeling}} We use this baseline for the text datasets only (BiasedSST and \founta{}).
    \item \textbf{P-DRO}: the parametric DRO approach proposed by \citet{michel2021modeling}. For image datasets, preliminary experiments using auto-regressive models for image modeling \citep{van2016pixel} proved to be prohibitively slow. Therefore, we only report P-DRO on text datasets as in \citet{michel2021modeling}.
    \item \textbf{Oracle DRO}: an online algorithm for minimizing the worst-case loss on the true uncertainty set \citep{sagawa2019distributionally}. Contrary to all other methods, Oracle DRO presupposes that we know the groups of interest at training time. As such, it is not directly comparable and serves to provide an upper bound of the robust accuracy that can be achieved.
\end{itemize}

Across all experiments, we report results with mean and standard deviation across 5 reruns with different seeds. Note that the model selection criterion can have a large effect on final performance \citep{gulrajani2020search}. In contrast to some related work (\textit{e.g.} \citet{koh2020wilds,idrissi2021simple}), except for Oracle DRO we only use validation criteria that \emph{do not} require group annotation on the validation set. We provide more details on the various hyper-parameters used in Appendix \ref{sec:appendix_details}.

\subsection{Results}

For all methods, we report the worst accuracy over all groups in the test set (the ``robust accuracy''). Models that are robust to subpopulation shift will have higher robust accuracies. Since there is often a trade-off between being robust to distribution shift and performing well on the training distribution, we also report the standard accuracy on the original test set (the ``average accuracy'')

As shown in Table \ref{tab:results_text}, RP-DRO works particularly well on the two text classification tasks, beating all baselines by a large margin on both BiasedSST and \founta{}. In fact, its robust accuracy almost matches that of Oracle DRO on \founta{}, despite the fact that the former does not use an any group information at training time. Compared to P-DRO, we find that results are not only better, but also more consistent as evidenced by the lower standard deviation.

Results are also generally good on image datasets (see Table \ref{tab:results_images}). On Waterbirds both the NonParam baseline and RP-DRO perform much better than ERM, with a slight edge for RP-DRO (although the latter exhibits a higher variance across reruns). On CelebA, RP-DRO largely outperforms the baselines.

\vspace{-0.5em}
\section{Analysis and Ablations}

We perform a series of analysis experiments and ablation studies to better (1) identify how the parametric representation of the ratio provides improvements over nonparametric alternatives and (2) understand the importance of the various choices made with regards to the renormalization scheme described in Section \ref{sec:method}. Throughout this section, experiments are performed on the BiasedSST dataset.

\vspace{-0.5em}
\subsection{Why are Parametric Adversaries Better? The Case of Label Noise}
\label{sec:noise}

Our experimental results in Section \ref{sec:experiments} shows that parametric approaches such as P-DRO and RP-DRO consistently outperform their nonparametric counterparts. A possible explanation for this phenomenon is that for nonparametric approaches, optimal weights generally only depends on the loss of the model. This can be problematic because the nonparametric worst-case distribution will indiscriminately up-weight noisy samples that have high loss. On the other hand, we hypothesize that it is more difficult for the parametric adversary to ``fit to the noise'' and that it tends to focus more on systematic patterns of failures of the model.

To corroborate this hypothesis, we perform experiments by adding increasing amounts of noise to the BiasedSST. Specifically, for each example in the training set we replace the original label with a random label with probability $p_{\text{noise}}= 0, 0.1,\ldots,0.5$. We then train models on these increasingly noisy datasets using both a parametric (RP-DRO) and nonparametric (NonParam-\{KL, CVaR,$\chi^2$\}) approach. To simplify experiments we only run one hyper-parameter configuration for each ($\tau=0.1$ and $\kappa=1$ for RP-DRO and NonParam respectively) and report the test accuracies of the model with the highest robust accuracy on the validation set. Results are averaged over 5 runs with different random seeds.

\begin{figure}
\centering
\begin{minipage}{.45\textwidth}
    \centering
    \includegraphics[width=\columnwidth]{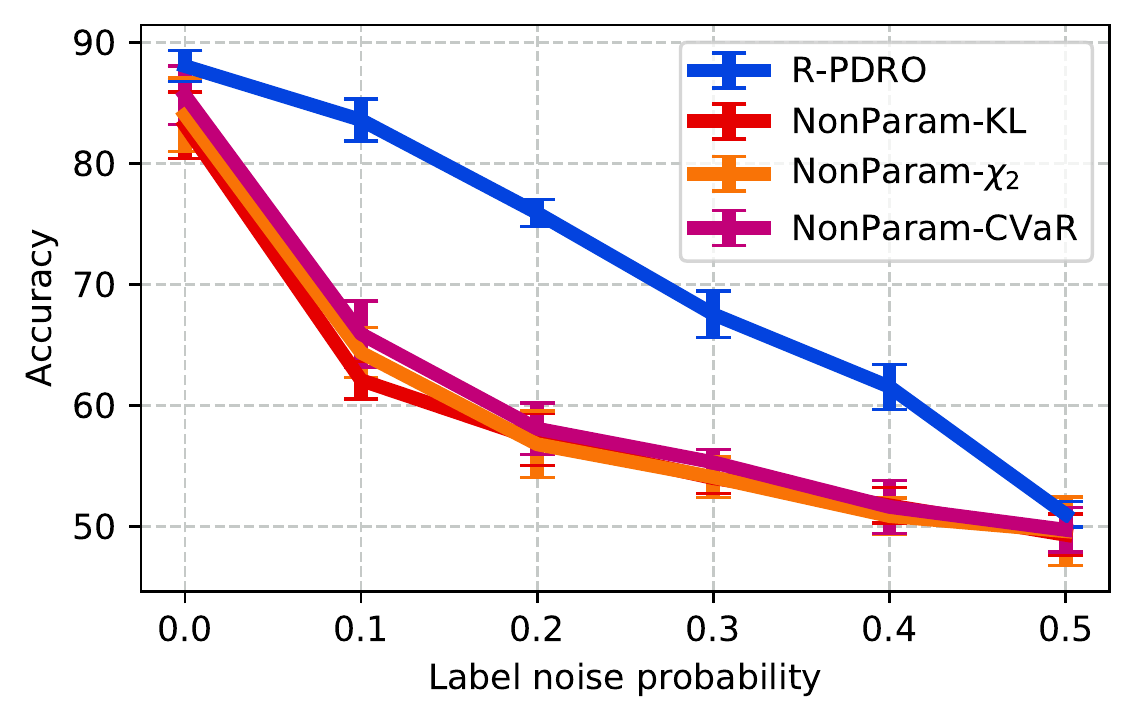}
    \caption{\label{fig:noise}Effect of label noise on performance (average test accuracy on BiasedSST) in parametric and nonparametric DRO.}
\end{minipage}~~
\begin{minipage}{.45\textwidth}
    \centering
    \includegraphics[width=\columnwidth]{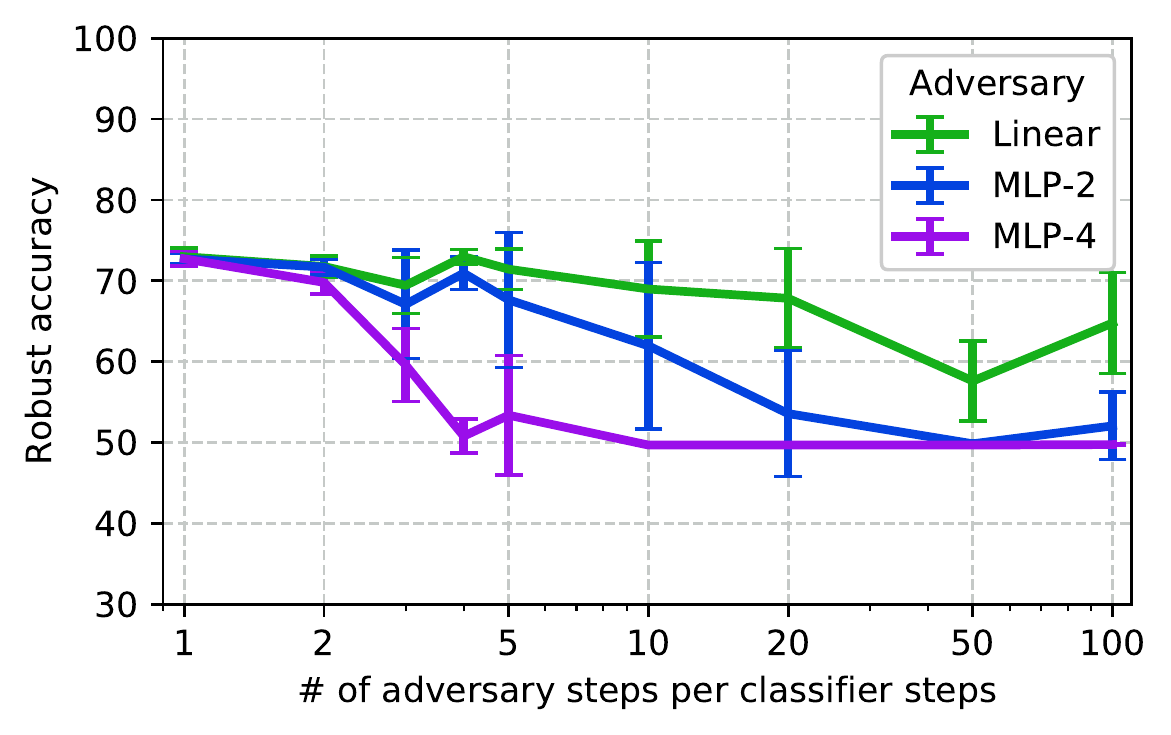}
    \caption{\label{fig:adv_steps}Evolution of RP-DRO's robust accuracy as the adversary takes more gradient steps than the classifier in a toy setting.}
\end{minipage}
\vspace{-2em}
\end{figure}

As showcased in Figure \ref{fig:noise}, we find that nonparametric approaches are very sensitive to label noise, losing around $20$ points when adding the smallest amount of noise ($p_{\text{noise}}=0.1$ ($85\rightarrow 65$), whereas RP-DRO is comparatively more robust with only a loss of around $5$ points for the same amount of noise. The trend holds all the way to $p_{\text{noise}}=0.5$ where all models collapse to chance accuracies. This further supports our hypothesis that nonparametric adversaries tend to fit to these noisy examples, which decreases the overall quality of the resulting classifier. We show a finer-grained analysis of the evolution of accuracies broken down by groups in Appendix \ref{sec:appendix_noise}. In Appendix \ref{sec:appendix_qualitative}, we present additional qualitative evidence that nonparametric approaches tends to select individually difficult examples rather than difficult subpopulations.

\vspace{-0.5em}
\subsection{Optimization with Simultaneous Gradient Updates Plays a Crucial Role}
\label{sec:simultaneous_gd}

Despite the aforementionned results, it remains unclear \emph{why} RP-DRO learns re-weightings that are less sensitive to label noise or difficult examples compared to NonParam-* methods. Indeed, since the nonparametric adversary is the optimal solution of the inner minimization problem in Eq. \ref{eq:ratio_dro_problem}, it stands to reason that (large, sometimes over-parameterized) parametric adversaries from RP-DRO would converge a solution close to NonParam. Our hypothesis is that the simultaneous updates to both model and adversary parameters prevent the parametric adversary from converging towards such solutions, and provides some implicit regularization against up-weighting examples that are noisy or too difficult.

To verify this hypothesis, we conduct a toy experiment where we allow the adversary to take additional gradient steps in-between each update to the classifier. At the limit, this would allow the adversary to find an optimum of the inner maximization problem at each step of training the classifier (which for large enough adversaries, might come close to the nonparametric solution).

For computational efficiency, these experiments are performed on a toy setting similar to that of \citet{michel2021modeling}: a linear model is trained on a binary classification problem with two domains, one of which is severely under-represented. For our adversary, we experiment with a linear adversary, as well as larger multilayer perceptrons with one hidden layer and 2 (MLP-2) and 4 (MLP-4) hidden units. In Figure \ref{fig:adv_steps}, we report the average robust accuracy (across 5 reruns) for classifiers trained with RP-DRO when the adversary is allowed to take more steps than the classifier.

We observe that RP-DRO’s robust accuracy suffers from giving the adversary too much time to catch up with the classifier: as the number of updates to the adversary increases, robust accuracy decreases. This effect is amplified in larger adversaries (\eg{} MLP-4). We find that a key effect of simultaneous updates is to dramatically improve stability (see Appendix \ref{sec:simultaneous_gd_stability}). This experiment underlines the importance of the simultaneous gradient updates, which prevent large, over-parameterized adversaries from converging to the sub-optimal nonparametric solution. We also find that simultaneous updates lead to dramatic

\vspace{-0.5em}
\subsection{Batch-level Normalization vs Self-normalization}

\begin{figure}[!t]
\centering
\includegraphics[width=0.85\columnwidth]{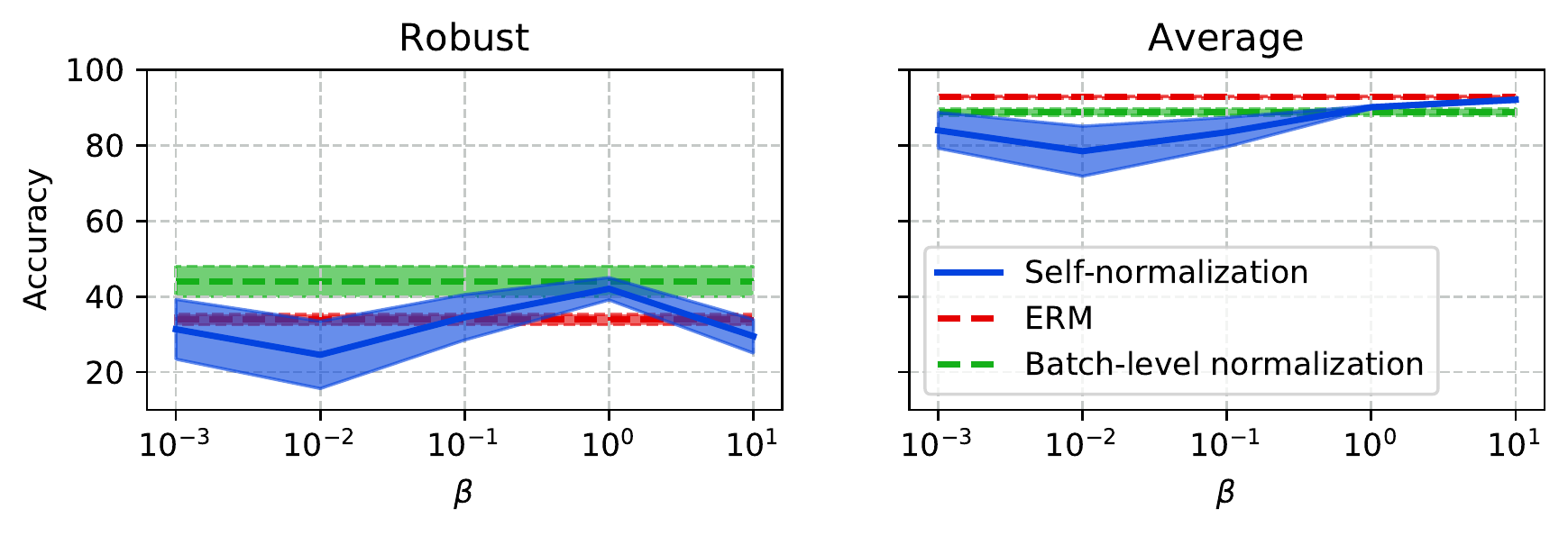}
\caption{\label{fig:self_norm_minmax} Effect of self-normalization coefficient $\beta$ on robust and average accuracy. We report results of ERM (which corresponds to $\beta=\infty$) and batch level renormalization for comparison.}
\vspace{-1.5em}
\end{figure}

In Section \ref{sec:normalization}, we discussed two alternatives for enforcing the normalization constraint on the ratios ($\mathbb E_{p}r_\psi=1$): a regularization-based approach (``self-normalization'') and batch level renormalization. In Figure \ref{fig:self_norm_minmax}, we show the effect of self-normalization with different values of the regularization weight $\beta$ for a fixed value of the KL penalty $\tau=0.1$. We find that batch-level normalization achieves a slightly better robust accuracy than the best self-normalization configuration.

In \citet{michel2021modeling}, the Minmax stopping criterion was identified as a major contributing factor to the performance of parametric DRO approaches. To understand how it affects each normalization strategy, we perform the same experiment as above but this time \emph{without} the Minmax criterion, selecting instead the model with the highest robust accuracy on the validation set (``Oracle stopping''), which provides an upper bound to the robust accuracy that can be achieved. We find that although accuracies are generally closer, batch-level normalization matches the best self-normalization penalty (the full figure can be found in Appendix \ref{sec:self_norm_oracle}). This indicates that batch-level normalization not only performs as well as self-normalization (without the need for tuning the additional hyper-parameter $\beta$), but also that it interacts better with the Minmax stopping criterion, making it a preferable alternative.

\vspace{-1em}
\subsection{Effect of Batch Size on Re-normalization}

\begin{wraptable}{r}{0.4\textwidth}
\vspace{-1.5em}
\begin{center}
\small
\caption{\label{tab:bsz_ablations} Effect of batch size on RP-DRO performance.}
\begin{tabular}{rrr}
\toprule
Batch size & Robust & Average \\
\midrule
64 (ERM) & { 32.97}  {\scriptsize$\pm$ 2.34} & { 92.42}  {\scriptsize$\pm$ 0.38} \\
\midrule
4 & { 37.50}  {\scriptsize$\pm$ 2.37} & { 92.32}  {\scriptsize$\pm$ 0.75} \\
8 & { 38.08}  {\scriptsize$\pm$ 2.06} & { 91.61}  {\scriptsize$\pm$ 0.48} \\
16 &  41.67  {\scriptsize$\pm$ 3.53} & {91.24}  {\scriptsize$\pm$ 0.34} \\
32 & 42.32  {\scriptsize$\pm$ 0.72} & { 89.77}  {\scriptsize$\pm$ 1.42} \\
64 & { 44.15}  {\scriptsize$\pm$ 6.83} & { 88.30}  {\scriptsize$\pm$ 2.33} \\
128 & { 42.25}  {\scriptsize$\pm$ 7.90} & { 88.21}  {\scriptsize$\pm$ 2.02} \\
\bottomrule
\end{tabular}
\end{center}
\vspace{-2em}
\end{wraptable}

As pointed out in Section \ref{sec:normalization}, a potential downside of the mini-batch-level normalization approach is that the effective weight of each sample then depends on the weights of the other samples within the mini-batch. For example, consider an adversary that assigns high weight to only 5\% of the training data. With a small enough batch size, it is likely that some batches may not contain any example of the high weight subpopulation, in which case mini-batch level renormalization will overestimate the weight of the sample in the mini-batch.

To assess the severity of this issue, we run RP-DRO on BiasedSST with $\tau=0.1$ and vary the batch size in $\{4, 8, 16, 32, 64, 128\}$. Each configuration is run 3 times, and we report average and standard deviation of the robust and average test accuracies in Table \ref{tab:bsz_ablations}. Results suggest that while robust accuracy indeed deteriorates for lower batch sizes (4 and 8), results are consistently good for batch sizes upwards of 16, a reasonable number considering that larger batch sizes are often preferred in the literature \citep{popel2018training,goyal2017accurate}.

\vspace{-0.5em}
\section{Conclusion}

In this paper we have proposed a parametric, likelihood ratio based approach to distributionally robust optimization of machine learning models. With the proposed method, we can use any type of parametric function estimator to define the uncertainty set of the DRO min-max game. We showed that with a careful renormalization strategy, the proposed method (RP-DRO) can be used to train robust models. It depends on very few hyper-parameters and consistently performs well on a number of benchmarks, making it an appealing ``off-the-shelf'' option. Finally we have shown that such parametric approaches are more resilient to the presence of noise in the training data when compared to their nonparametric alternatives, and that simultaneous gradient descent is a key component of RP-DRO's success.

The main downside of RP-DRO is the computational overhead of jointly training a second neural model. An interesting direction for future work is to improve its efficiency through parallel computation or by sharing parameters between the classifier and the adversary.

\section*{Acknowledgments}

The authors would like to thank the anonymous reviewers whose feedback helped improve the paper to its current form. Moreover, this project benefited from fruitful discussions with Chunting Zhou, Daniel Levy, Zachary Lipton and Zico Kolter. The first author was supported by the ENS-CFM Data Science Chair.

\section*{Ethics Statement}

The work presented in this paper places itself in the continuity of series of papers dedicated to training models that are more robust to certain types of distribution shifts (\citet{sagawa2019distributionally,levy2020large,zhou2021examining} to cite only a few). This line of work allows for training models that perform more equitably across subpopulations of the training set, with positive implications in terms of fairness and equal representations of marginalized groups.

A potential negative impact of RP-DRO is in the increased computational overhead that results from training the adversary, which leads to a larger carbon footprint, and ultimately negatively affects the environment. This is a shortcoming which we hope to address in future work.

\section*{Reproducibility Statement}

To facilitate reproducibility of our work, our experiments are conducted on datasets that are openly available: BiasedSST\footnote{Original dataset found at \url{https://nlp.stanford.edu/sentiment/index.html}, code to create the biased version can be found in \citet{michel2021modeling}.}, \founta{}\footnote{Original dataset found at \url{http://ow.ly/BqCf30jqffN}, instructions to obtain automatic dialect annotation found in \citet{blodgett2016demographic}.} \citep{founta2018large}, Waterbirds\footnote{\url{https://nlp.stanford.edu/data/dro/waterbird_complete95_forest2water2.tar.gz}} \citep{sagawa2019distributionally} and CelebA\footnote{\url{http://mmlab.ie.cuhk.edu.hk/projects/CelebA.html}} \citep{liu2015faceattributes}. We describe experimental settings in Section \ref{sec:expt_setting}, and additional hyper-parameters are reported in Appendix \ref{sec:appendix_details}. Code to reproduce our experiments is available at \codeurl{}.

\bibliography{bibliography/myplain.bib,bibliography/references.bib}

\begin{thebibliography}{50}
\providecommand{\natexlab}[1]{#1}
\providecommand{\url}[1]{\texttt{#1}}
\expandafter\ifx\csname urlstyle\endcsname\relax
  \providecommand{\doi}[1]{doi: #1}\else
  \providecommand{\doi}{doi: \begingroup \urlstyle{rm}\Url}\fi

\bibitem[Amodei et~al.(2016)Amodei, Ananthanarayanan, Anubhai, Bai, Battenberg,
  Case, Casper, Catanzaro, Cheng, Chen, Chen, Chen, Chen, Chrzanowski, Coates,
  Diamos, Ding, Du, Elsen, Engel, Fang, Fan, Fougner, Gao, Gong, Hannun, Han,
  Johannes, Jiang, Ju, Jun, LeGresley, Lin, Liu, Liu, Li, Li, Ma, Narang, Ng,
  Ozair, Peng, Prenger, Qian, Quan, Raiman, Rao, Satheesh, Seetapun, Sengupta,
  Srinet, Sriram, Tang, Tang, Wang, Wang, Wang, Wang, Wang, Wang, Wu, Wei,
  Xiao, Xie, Xie, Yogatama, Yuan, Zhan, and Zhu]{amodei2016deep}
Dario Amodei, Sundaram Ananthanarayanan, Rishita Anubhai, Jingliang Bai, Eric
  Battenberg, Carl Case, Jared Casper, Bryan Catanzaro, Qiang Cheng, Guoliang
  Chen, Jie Chen, Jingdong Chen, Zhijie Chen, Mike Chrzanowski, Adam Coates,
  Greg Diamos, Ke~Ding, Niandong Du, Erich Elsen, Jesse Engel, Weiwei Fang,
  Linxi Fan, Christopher Fougner, Liang Gao, Caixia Gong, Awni Hannun, Tony
  Han, Lappi Johannes, Bing Jiang, Cai Ju, Billy Jun, Patrick LeGresley, Libby
  Lin, Junjie Liu, Yang Liu, Weigao Li, Xiangang Li, Dongpeng Ma, Sharan
  Narang, Andrew Ng, Sherjil Ozair, Yiping Peng, Ryan Prenger, Sheng Qian,
  Zongfeng Quan, Jonathan Raiman, Vinay Rao, Sanjeev Satheesh, David Seetapun,
  Shubho Sengupta, Kavya Srinet, Anuroop Sriram, Haiyuan Tang, Liliang Tang,
  Chong Wang, Jidong Wang, Kaifu Wang, Yi~Wang, Zhijian Wang, Zhiqian Wang,
  Shuang Wu, Likai Wei, Bo~Xiao, Wen Xie, Yan Xie, Dani Yogatama, Bin Yuan, Jun
  Zhan, and Zhenyao Zhu.
\newblock Deep speech 2 : End-to-end speech recognition in english and
  mandarin.
\newblock In \emph{Proceedings of the 33rd International Conference on Machine
  Learning (ICML)}, pp.\  173--182, 2016.
\newblock URL \url{http://proceedings.mlr.press/v48/amodei16.html}.

\bibitem[Andreas et~al.(2015)Andreas, Rabinovich, Jordan, and
  Klein]{andreas2015accuracy}
Jacob Andreas, Maxim Rabinovich, Michael~I Jordan, and Dan Klein.
\newblock On the accuracy of self-normalized log-linear models.
\newblock In \emph{Proceedings of the 29th Annual Conference on Neural
  Information Processing Systems (NIPS)}, pp.\  1783--1791, 2015.

\bibitem[Balduzzi et~al.(2018)Balduzzi, Racaniere, Martens, Foerster, Tuyls,
  and Graepel]{balduzzi2018mechanics}
David Balduzzi, Sebastien Racaniere, James Martens, Jakob Foerster, Karl Tuyls,
  and Thore Graepel.
\newblock The mechanics of n-player differentiable games.
\newblock In \emph{Proceedings of the 35th International Conference on Machine
  Learning (ICML)}, pp.\  354--363, 2018.
\newblock URL
  \url{http://proceedings.mlr.press/v80/balduzzi18a/balduzzi18a.pdf}.

\bibitem[Beery et~al.(2018)Beery, Van~Horn, and Perona]{beery2018recognition}
Sara Beery, Grant Van~Horn, and Pietro Perona.
\newblock Recognition in terra incognita.
\newblock In \emph{Proceedings of the 16th European Conference on Computer
  Vision (ECCV)}, pp.\  456--473, 2018.

\bibitem[Ben-Tal et~al.(2013)Ben-Tal, Den~Hertog, De~Waegenaere, Melenberg, and
  Rennen]{ben2013robust}
Aharon Ben-Tal, Dick Den~Hertog, Anja De~Waegenaere, Bertrand Melenberg, and
  Gijs Rennen.
\newblock Robust solutions of optimization problems affected by uncertain
  probabilities.
\newblock \emph{Management Science}, 59\penalty0 (2):\penalty0 341--357, 2013.

\bibitem[Blodgett et~al.(2016)Blodgett, Green, and
  O{'}Connor]{blodgett2016demographic}
Su~Lin Blodgett, Lisa Green, and Brendan O{'}Connor.
\newblock Demographic dialectal variation in social media: A case study of
  {A}frican-{A}merican {E}nglish.
\newblock In \emph{Proceedings of the 54th Annual Meeting of the Association
  for Computational Linguistics (ACL)}, pp.\  1119--1130, 2016.
\newblock URL \url{https://www.aclweb.org/anthology/D16-1120}.

\bibitem[Buolamwini \& Gebru(2018)Buolamwini and Gebru]{buolamwini2018gender}
Joy Buolamwini and Timnit Gebru.
\newblock Gender shades: Intersectional accuracy disparities in commercial
  gender classification.
\newblock In \emph{Conference on fairness, accountability and transparency},
  pp.\  77--91. PMLR, 2018.

\bibitem[Cortes et~al.(2010)Cortes, Mansour, and Mohri]{cortes2010learning}
Corinna Cortes, Yishay Mansour, and Mehryar Mohri.
\newblock Learning bounds for importance weighting.
\newblock In \emph{Proceedings of the 23rd Annual Conference on Neural
  Information Processing Systems (NIPS)}, volume~10, pp.\  442--450. Citeseer,
  2010.

\bibitem[Curi et~al.(2020)Curi, Levy, Jegelka, and Krause]{curi2020adaptive}
Sebastian Curi, Kfir~Y. Levy, Stefanie Jegelka, and Andreas Krause.
\newblock Adaptive sampling for stochastic risk-averse learning.
\newblock In H.~Larochelle, M.~Ranzato, R.~Hadsell, M.~F. Balcan, and H.~Lin
  (eds.), \emph{Proceedings of the 34th Annual Conference on Neural Information
  Processing Systems (NeurIPS)}, volume~33, pp.\  1036--1047. Curran
  Associates, Inc., 2020.
\newblock URL
  \url{https://proceedings.neurips.cc/paper/2020/file/0b6ace9e8971cf36f1782aa982a708db-Paper.pdf}.

\bibitem[Delage \& Ye(2010)Delage and Ye]{delage2010distributionally}
Erick Delage and Yinyu Ye.
\newblock Distributionally robust optimization under moment uncertainty with
  application to data-driven problems.
\newblock \emph{Operations research}, 58\penalty0 (3):\penalty0 595--612, 2010.

\bibitem[Deng et~al.(2009)Deng, Dong, Socher, Li, Li, and
  Fei-Fei]{deng2009imagenet}
Jia Deng, Wei Dong, Richard Socher, Li-Jia Li, Kai Li, and Li~Fei-Fei.
\newblock Imagenet: A large-scale hierarchical image database.
\newblock In \emph{Proceedings of the 22nd IEEE Conference on Computer Vision
  and Pattern Recognition (CVPR)}, pp.\  248--255, 2009.

\bibitem[Devlin et~al.(2018)Devlin, Chang, Lee, and Toutanova]{devlin2018bert}
Jacob Devlin, Ming-Wei Chang, Kenton Lee, and Kristina Toutanova.
\newblock Bert: Pre-training of deep bidirectional transformers for language
  understanding.
\newblock In \emph{Proceedings of the 2019 Conference of the North American
  Chapter of the Association for Computational Linguistics: Human Language
  Technologies (NAACL-HLT)}, 2018.

\bibitem[Duchi et~al.(2020)Duchi, Hashimoto, and Namkoong]{DuHa20dis}
J.~Duchi, T.~Hashimoto, and H.~Namkoong.
\newblock Distributionally robust losses for latent covariate mixtures.
\newblock \emph{arXiv preprint arXiv:2007.13982}, 2020.

\bibitem[Duchi \& Namkoong(2018)Duchi and Namkoong]{duchi2018learning}
John Duchi and Hongseok Namkoong.
\newblock Learning models with uniform performance via distributionally robust
  optimization.
\newblock \emph{arXiv preprint arXiv:1810.08750}, 2018.
\newblock URL \url{https://arxiv.org/pdf/1810.08750.pdf}.

\bibitem[Fan et~al.(2017)Fan, Lyu, Ying, and Hu]{fan2017learning}
Yanbo Fan, Siwei Lyu, Yiming Ying, and Bao-Gang Hu.
\newblock Learning with average top-k loss.
\newblock In \emph{Proceedings of the 31st Annual Conference on Neural
  Information Processing Systems (NIPS)}, pp.\  497--505, 2017.

\bibitem[Faury et~al.(2020)Faury, Tanielian, Dohmatob, Smirnova, and
  Vasile]{faury2020distributionally}
Louis Faury, Ugo Tanielian, Elvis Dohmatob, Elena Smirnova, and Flavian Vasile.
\newblock Distributionally robust counterfactual risk minimization.
\newblock In \emph{Proceedings of the 34th Meeting of the Association for
  Advancement of Artificial Intelligence (AAAI)}, volume~34, pp.\  3850--3857,
  2020.

\bibitem[Founta et~al.(2018)Founta, Djouvas, Chatzakou, Leontiadis, Blackburn,
  Stringhini, Vakali, Sirivianos, and Kourtellis]{founta2018large}
Antigoni-Maria Founta, Constantinos Djouvas, Despoina Chatzakou, Ilias
  Leontiadis, Jeremy Blackburn, Gianluca Stringhini, Athena Vakali, Michael
  Sirivianos, and Nicolas Kourtellis.
\newblock Large scale crowdsourcing and characterization of twitter abusive
  behavior.
\newblock In \emph{Proceedings of the 12th International AAAI Conference on
  Weblogs and Social Media (ICWSM)}, 2018.

\bibitem[Goyal et~al.(2019)Goyal, Dyer, and
  Berg-Kirkpatrick]{goyal2019empirical}
Kartik Goyal, Chris Dyer, and Taylor Berg-Kirkpatrick.
\newblock An empirical investigation of global and local normalization for
  recurrent neural sequence models using a continuous relaxation to beam
  search.
\newblock In \emph{Proceedings of the 2019 Conference of the North American
  Chapter of the Association for Computational Linguistics: Human Language
  Technologies (NAACL-HLT)}, pp.\  1724--1733, 2019.

\bibitem[Goyal et~al.(2017)Goyal, Doll{\'a}r, Girshick, Noordhuis, Wesolowski,
  Kyrola, Tulloch, Jia, and He]{goyal2017accurate}
Priya Goyal, Piotr Doll{\'a}r, Ross Girshick, Pieter Noordhuis, Lukasz
  Wesolowski, Aapo Kyrola, Andrew Tulloch, Yangqing Jia, and Kaiming He.
\newblock Accurate, large minibatch sgd: Training imagenet in 1 hour.
\newblock \emph{arXiv preprint arXiv:1706.02677}, 2017.

\bibitem[Grother et~al.(2019)Grother, Ngan, and Hanaoka]{grother2019face}
Patrick Grother, Mei Ngan, and Kayee Hanaoka.
\newblock \emph{Face Recognition Vendor Test (FVRT): Part 3, Demographic
  Effects}.
\newblock National Institute of Standards and Technology, 2019.

\bibitem[Gulrajani \& Lopez-Paz(2020)Gulrajani and
  Lopez-Paz]{gulrajani2020search}
Ishaan Gulrajani and David Lopez-Paz.
\newblock In search of lost domain generalization.
\newblock In \emph{Proceedings of the International Conference on Learning
  Representations (ICLR)}, 2020.

\bibitem[Gururangan et~al.(2020)Gururangan, Marasovi{\'c}, Swayamdipta, Lo,
  Beltagy, Downey, and Smith]{gururangan-etal-2020-dont}
Suchin Gururangan, Ana Marasovi{\'c}, Swabha Swayamdipta, Kyle Lo, Iz~Beltagy,
  Doug Downey, and Noah~A. Smith.
\newblock Don{'}t stop pretraining: Adapt language models to domains and tasks.
\newblock In \emph{Proceedings of the 58th Annual Meeting of the Association
  for Computational Linguistics}, pp.\  8342--8360, Online, July 2020.
  Association for Computational Linguistics.
\newblock \doi{10.18653/v1/2020.acl-main.740}.
\newblock URL \url{https://www.aclweb.org/anthology/2020.acl-main.740}.

\bibitem[Hashimoto et~al.(2018)Hashimoto, Srivastava, Namkoong, and
  Liang]{hashimoto2018fairness}
Tatsunori Hashimoto, Megha Srivastava, Hongseok Namkoong, and Percy Liang.
\newblock Fairness without demographics in repeated loss minimization.
\newblock In \emph{Proceedings of the 35th International Conference on Machine
  Learning (ICML)}, pp.\  1929--1938. PMLR, 2018.

\bibitem[He et~al.(2016)He, Zhang, Ren, and Sun]{he2016deep}
Kaiming He, Xiangyu Zhang, Shaoqing Ren, and Jian Sun.
\newblock Deep residual learning for image recognition.
\newblock In \emph{Proceedings of the 29th IEEE Conference on Computer Vision
  and Pattern Recognition (CVPR)}, pp.\  770--778, 2016.

\bibitem[Hovy \& S{\o}gaard(2015)Hovy and S{\o}gaard]{hovy2015tagging}
Dirk Hovy and Anders S{\o}gaard.
\newblock Tagging performance correlates with author age.
\newblock In \emph{Proceedings of the 53rd Annual Meeting of the Association
  for Computational Linguistics (ACL)}, pp.\  483--488, 2015.
\newblock URL \url{https://www.aclweb.org/anthology/P15-2079}.

\bibitem[Hu et~al.(2018)Hu, Niu, Sato, and Sugiyama]{hu2018does}
Weihua Hu, Gang Niu, Issei Sato, and Masashi Sugiyama.
\newblock Does distributionally robust supervised learning give robust
  classifiers?
\newblock In \emph{Proceedings of the 35th International Conference on Machine
  Learning (ICML)}, pp.\  2029--2037, 2018.
\newblock URL \url{http://proceedings.mlr.press/v80/hu18a/hu18a.pdf}.

\bibitem[Hu \& Hong(2013)Hu and Hong]{hu2013kullback}
Zhaolin Hu and L~Jeff Hong.
\newblock Kullback-leibler divergence constrained distributionally robust
  optimization.
\newblock \emph{Available at Optimization Online}, 2013.

\bibitem[Husain(2020)]{husain2020distributional}
Hisham Husain.
\newblock Distributional robustness with ipms and links to regularization and
  gans.
\newblock In \emph{Proceedings of the 34th Annual Conference on Neural
  Information Processing Systems (NeurIPS)}, 2020.

\bibitem[Idrissi et~al.(2021)Idrissi, Arjovsky, Pezeshki, and
  Lopez-Paz]{idrissi2021simple}
Badr~Youbi Idrissi, Martin Arjovsky, Mohammad Pezeshki, and David Lopez-Paz.
\newblock Simple data balancing achieves competitive worst-group-accuracy.
\newblock In \emph{Proceedings of the 2st Conference on Causal Learning and
  Reasoning (CLeaR)}, 2021.

\bibitem[Kingma \& Ba(2014)Kingma and Ba]{Kingma2014Adam}
Diederik~P. Kingma and Jimmy Ba.
\newblock Adam: A method for stochastic optimization.
\newblock In \emph{Proceedings of the International Conference on Learning
  Representations (ICLR)}, 2014.

\bibitem[Koh et~al.(2020)Koh, Sagawa, Marklund, Xie, Zhang, Balsubramani, Hu,
  Yasunaga, Phillips, Gao, et~al.]{koh2020wilds}
Pang~Wei Koh, Shiori Sagawa, Henrik Marklund, Sang~Michael Xie, Marvin Zhang,
  Akshay Balsubramani, Weihua Hu, Michihiro Yasunaga, Richard~Lanas Phillips,
  Irena Gao, et~al.
\newblock Wilds: A benchmark of in-the-wild distribution shifts.
\newblock \emph{arXiv preprint arXiv:2012.07421}, 2020.

\bibitem[Levy et~al.(2020)Levy, Carmon, Duchi, and Sidford]{levy2020large}
Daniel Levy, Yair Carmon, John~C Duchi, and Aaron Sidford.
\newblock Large-scale methods for distributionally robust optimization.
\newblock \emph{Proceedings of the 34th Annual Conference on Neural Information
  Processing Systems (NeurIPS)}, 33, 2020.

\bibitem[Liu et~al.(2015)Liu, Luo, Wang, and Tang]{liu2015faceattributes}
Ziwei Liu, Ping Luo, Xiaogang Wang, and Xiaoou Tang.
\newblock Deep learning face attributes in the wild.
\newblock In \emph{Proceedings of the 15th International Conference on Computer
  Vision (ICCV)}, December 2015.

\bibitem[McCoy et~al.(2019)McCoy, Pavlick, and Linzen]{mccoy2019right}
Tom McCoy, Ellie Pavlick, and Tal Linzen.
\newblock Right for the wrong reasons: Diagnosing syntactic heuristics in
  natural language inference.
\newblock In \emph{Proceedings of the 57th Annual Meeting of the Association
  for Computational Linguistics (ACL)}, pp.\  3428--3448, 2019.

\bibitem[Michel \& Neubig(2018)Michel and Neubig]{michel18mtnt}
Paul Michel and Graham Neubig.
\newblock {MTNT}: A testbed for machine translation of noisy text.
\newblock In \emph{Proceedings of the 2018 Conference on Empirical Methods in
  Natural Language Processing (EMNLP)}, pp.\  543--553, 2018.

\bibitem[Michel et~al.(2021)Michel, Hashimoto, and Neubig]{michel2021modeling}
Paul Michel, Tatsunori Hashimoto, and Graham Neubig.
\newblock Modeling the second player in distributionally robust optimization.
\newblock In \emph{Proceedings of the International Conference on Learning
  Representations (ICLR)}, 2021.

\bibitem[Nguyen et~al.(2020)Nguyen, Si, and Blanchet]{nguyen2020robust}
Viet~Anh Nguyen, Nian Si, and Jose Blanchet.
\newblock Robust bayesian classification using an optimistic score ratio.
\newblock In \emph{Proceedings of the 37th International Conference on Machine
  Learning (ICML)}, 2020.

\bibitem[Oren et~al.(2019)Oren, Sagawa, Hashimoto, and
  Liang]{oren2019distributionally}
Yonatan Oren, Shiori Sagawa, Tatsunori Hashimoto, and Percy Liang.
\newblock Distributionally robust language modeling.
\newblock In \emph{Proceedings of the 2019 Conference on Empirical Methods in
  Natural Language Processing (EMNLP)}, pp.\  4227--4237, 2019.
\newblock URL \url{https://www.aclweb.org/anthology/D19-1432}.

\bibitem[Popel \& Bojar(2018)Popel and Bojar]{popel2018training}
Martin Popel and Ond{\v{r}}ej Bojar.
\newblock Training tips for the transformer model.
\newblock \emph{arXiv preprint arXiv:1804.00247}, 2018.

\bibitem[Radford et~al.(2018)Radford, Narasimhan, Salimans, and
  Sutskever]{radford2018improving}
Alec Radford, Karthik Narasimhan, Time Salimans, and Ilya Sutskever.
\newblock Improving language understanding with unsupervised learning.
\newblock Technical report, Technical report, OpenAI, 2018.

\bibitem[Rahimian \& Mehrotra(2019)Rahimian and
  Mehrotra]{rahimian2019distributionally}
Hamed Rahimian and Sanjay Mehrotra.
\newblock Distributionally {R}obust {O}ptimization: A {R}eview.
\newblock \emph{arXiv preprint arXiv:1908.05659}, 2019.

\bibitem[Sagawa et~al.(2020)Sagawa, Koh, Hashimoto, and
  Liang]{sagawa2019distributionally}
Shiori Sagawa, Pang~Wei Koh, Tatsunori~B Hashimoto, and Percy Liang.
\newblock Distributionally robust neural networks for group shifts: On the
  importance of regularization for worst-case generalization.
\newblock In \emph{Proceedings of the International Conference on Learning
  Representations (ICLR)}, 2020.
\newblock URL \url{https://arxiv.org/pdf/1911.08731.pdf}.

\bibitem[Sanh et~al.(2019)Sanh, Debut, Chaumond, and Wolf]{sanh2019distilbert}
Victor Sanh, Lysandre Debut, Julien Chaumond, and Thomas Wolf.
\newblock Distilbert, a distilled version of bert: smaller, faster, cheaper and
  lighter.
\newblock \emph{arXiv preprint arXiv:1910.01108}, 2019.

\bibitem[Sap et~al.(2019)Sap, Card, Gabriel, Choi, and Smith]{sap2019risk}
Maarten Sap, Dallas Card, Saadia Gabriel, Yejin Choi, and Noah~A. Smith.
\newblock The risk of racial bias in hate speech detection.
\newblock In \emph{Proceedings of the 57th Annual Meeting of the Association
  for Computational Linguistics (ACL)}, pp.\  1668--1678, 2019.
\newblock URL \url{https://www.aclweb.org/anthology/P19-1163}.

\bibitem[Singh et~al.(2000)Singh, Kearns, and Mansour]{singh2000nash}
Satinder~P. Singh, Michael~J. Kearns, and Yishay Mansour.
\newblock Nash convergence of gradient dynamics in general-sum games.
\newblock In \emph{Proceedings of the 16th Conference on Uncertainty in
  Artificial Intelligence}, UAI '00, pp.\  541–548, San Francisco, CA, USA,
  2000. Morgan Kaufmann Publishers Inc.
\newblock ISBN 1558607099.

\bibitem[Sinha et~al.(2018)Sinha, Namkoong, and Duchi]{sinha2017certifying}
Aman Sinha, Hongseok Namkoong, and John Duchi.
\newblock Certifying some distributional robustness with principled adversarial
  training.
\newblock In \emph{Proceedings of the International Conference on Learning
  Representations (ICLR)}, 2018.

\bibitem[Van~Oord et~al.(2016)Van~Oord, Kalchbrenner, and
  Kavukcuoglu]{van2016pixel}
Aaron Van~Oord, Nal Kalchbrenner, and Koray Kavukcuoglu.
\newblock Pixel recurrent neural networks.
\newblock In \emph{Proceedings of the 33rd International Conference on Machine
  Learning (ICML)}, pp.\  1747--1756. PMLR, 2016.

\bibitem[Vapnik(1992)]{vapnik1992principles}
Vladimir Vapnik.
\newblock Principles of risk minimization for learning theory.
\newblock In \emph{Proceedings of the 5th Annual Conference on Neural
  Information Processing Systems (NIPS)}, pp.\  831--838, 1992.

\bibitem[Zech et~al.(2018)Zech, Badgeley, Liu, Costa, Titano, and
  Oermann]{zech2018variable}
John~R Zech, Marcus~A Badgeley, Manway Liu, Anthony~B Costa, Joseph~J Titano,
  and Eric~Karl Oermann.
\newblock Variable generalization performance of a deep learning model to
  detect pneumonia in chest radiographs: a cross-sectional study.
\newblock \emph{PLoS medicine}, 15\penalty0 (11):\penalty0 e1002683, 2018.

\bibitem[Zhou et~al.(2021)Zhou, Ma, Michel, and Neubig]{zhou2021examining}
Chunting Zhou, Xuezhe Ma, Paul Michel, and Graham Neubig.
\newblock Examining and combating spurious features under distribution shift.
\newblock In \emph{Proceedings of the 38th International Conference on Machine
  Learning (ICML)}, 2021.

\end{thebibliography}
\bibliographystyle{iclr2022_conference}

\clearpage

\appendix

\section{Additional Experimental Details}
\label{sec:appendix_details}

We describe additional hyper-parameter choices specific to each dataset or method to facilitate reproduction of our results.

\subsection{Dataset-specific Hyper-parameters}

All hyper-parameters listed below are constant across all methods:

\paragraph{Text Datasets} The input data is tokenized using the \texttt{bert-base-uncased} sub-word tokenizer from \citet{devlin2018bert}. We train both classifier and adversary with Adam \citep{Kingma2014Adam} using a learning rate of $2\times 10^{-5}$, linearly decay the learning rate to 0 at each step. We train with batches of size 64 (or containing up to 2500 tokens, whichever is lower) for 50 and 20 epochs for BiasedSST and \founta{} respectively, evaluating model on the validation data every epoch.

\paragraph{Image Datasets} On both datasets, images are rescaled to $224\times224$ pixels and pixel values are normalized to have mean 0 and variance 1 across all 3 color channels on the training data. At training time, we augment the data by randomly cropping or flipping the images horizontally. We train using regular stochastic gradient descent using a constant learning rate of $10^{-3}$ and a batch size of 32. We train for 75 and 13 epochs on Waterbirds and CelebA respectively (those numbers were chosen to match the number of steps trained to \citet{sagawa2019distributionally} despite the smaller batch size), and validate every 100 (for Waterbirds) and 1000 (for CelebA) training steps.

\subsection{Method Specific Hyper-parameters}

For NonParam we follow the adaptation of \citet{hu2018does} used in \citet{michel2021modeling} and choose the optimal temperature $\tau^*$ based on mini-batch level estimates of the KL divergence. We treat the KL bound $\kappa$ as a hyper-parameter. We adapt the Minmax stopping criterion of P-DRO to the nonparametric adversaries as we found it yielded more robust models than those selected with average validation accuracy. We sweep over $\kappa\in\{0.01,0.1,1.0,10.0\}$

For RP-DRO we perform min-max stopping using the Minmax criterion with a KL threshold of $\log 10$ in all experiments, to match the value recommended for P-DRO. Specifically, we estimate the KL divergence of checkpointed adversaries $\psi_i$ on the validation data as follows:
\begin{equation}
    \frac 1 {|\mathcal{D}_{\text{valid}}|}\sum_{x,y\in \mathcal{D}_{\text{valid}}}\hat{r}_\psi(x, y)\log \hat{r}_\psi(x, y)
\end{equation}
and reject adversaries for which this quantity exceeds $\log 10$.

\section{Additional Comparisons Between Parametric and nonparametric DRO}

\subsection{Evolution of Group Accuracies under Increasing Label Noise}
\label{sec:appendix_noise}

\begin{figure}[ht!]
    \centering
    \includegraphics[width=0.95\textwidth]{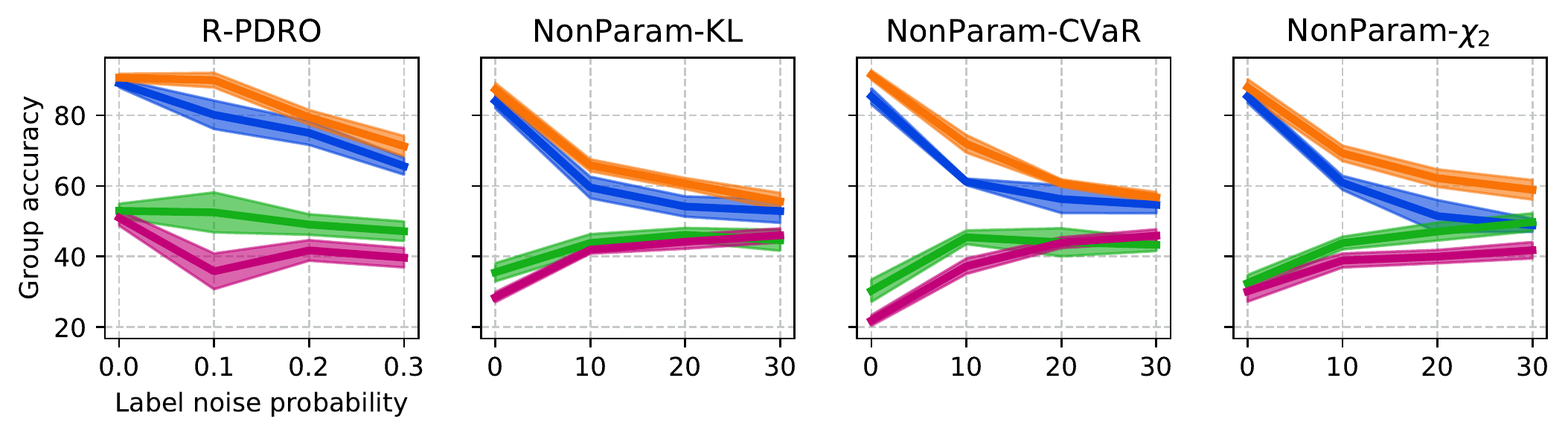}
    \caption{\label{fig:noise_groups}Effect of label noise on model performance on each individual group in parametric (R-PDRO) and non parametric DRO (NonParam-\{KL,$\chi_2$,CVaR\}).}
    \vspace{-1em}
\end{figure}

In Figure \ref{fig:noise_groups} we report the evolution of individual group accuracies under increasing amounts of label noise (as described in Section \ref{sec:noise}). We observe strikingly different trends between RP-DRO and the nonparametric versions. Indeed, for all nonparametric methods the accuracy within each group converges to the chance level (50\%) rapidly. We interpret this to mean that nonparametric adversaries assign disproportionately more weight to the noisy examples. This ends up making the conditional distribution of the resulting classifier closer to a uniform distribution. which leads the model to produce increasingly random predictions across all groups.

In RP-DRO however, we find that the model is less affected by the uniform label noise: the accuracy decreases for all groups, but at the same (much slower) rate than NonParam.

\subsection{Qualitative Comparison of Parametric and nonparametric Ratios}
\label{sec:appendix_qualitative}

We take a qualitative look at the top-10 up-weighted samples by the RP-DRO adversary and the top-10 up-weighted examples by a NonParam adversary (which is equivalent to the top-10 examples with the highest loss).

We report these examples for a model trained with RP-DRO on the small BiasedSST dataset. Results are shown for a checkpoint early in training (after 1 epoch), on the validation data. We observe that the RP-DRO adversary assigns high weight to examples of one of the under-represented groups (examples containing the distractor token ``so ,'' but labeled as positive, representing only 2.5\% of the training data).

\begin{table}[t!]
    \centering
    \begin{tabular}{c|p{0.8\columnwidth}}
    \toprule
         Label & Text\\
         \midrule
         \multicolumn{2}{c}{RP-DRO}\\
        \midrule
        positive & `` so, not only is undercover brother as funny, if not more so, than both austin powers films, but it's also one of the smarter, savvier spoofs to come along in some time. '' \\
        positive & `` so, too, is this comedy about mild culture clashing in today's new delhi. '' \\
        positive & `` so, we root for ( clara and paul ), even like them, though perhaps it's an emotion closer to pity. '' \\
        positive & `` so, thanks to scott's charismatic roger and eisenberg's sweet nephew, roger dodger is one of the most compelling variations on in the company of men. '' \\
        positive & `` so, the sort of film that makes me miss hitchcock, but also feel optimistic that there's hope for popular cinema yet. '' \\
        positive & `` so, visually imaginative, thematically instructive and thoroughly delightful, it takes us on a roller - coaster ride from innocence to experience without even a hint of that typical kiddie - flick sentimentality. '' \\
        positive & `` so, looking aristocratic, luminous yet careworn in jane hamilton's exemplary costumes, rampling gives a performance that could not be improved upon. ' '' \\
        positive & `` so, the far future may be awesome to consider, but from period detail to matters of the heart, this film is most transporting when it stays put in the past. '' \\
        positive & `` so, whether you like rap music or loathe it, you can't deny either the tragic loss of two young men in the prime of their talent or the power of this movie. '' \\
        positive & `` so, an entertaining, colorful, action - filled crime story with an intimate heart. '' \\
        \midrule
        \multicolumn{2}{c}{NonParam}\\
        \midrule
        % %--------------------------------------------------------------------------------
        % % Top-5 losses
        negative & `` that's a cheat. '' \\
        negative & `` its well of thorn and vinegar ( and simple humanity ) has long been plundered by similar works featuring the insight and punch this picture so conspicuously lacks. '' \\
        negative & `` so, an absurdist comedy about alienation, separation and loss. '' \\
        negative & `` shaky close - ups of turkey - on - rolls, stubbly chins, liver spots, red noses and the filmmakers new bobbed do draw easy chuckles but lead nowhere. '' \\
        negative & `` paid in full is so stale, in fact, that its most vibrant scene is one that uses clips from brian de palma's scarface. '' \\
        negative & `` so, and the lesson, in the end, is nothing new. '' \\
        negative & `` may reawaken discussion of the kennedy assassination but this fictional film looks made for cable rather than for the big screen. '' \\
        positive & `` so, atom egoyan has conjured up a multilayered work that tackles any number of fascinating issues '' \\
        negative & `` so, dull, lifeless, and amateurishly assembled. '' \\
        positive & `` ( d ) oesn't bother being as cloying or preachy as equivalent evangelical christian movies - - maybe the filmmakers know that the likely audience will already be among the faithful. '' \\
        \bottomrule
    \end{tabular}
    \caption{Top-10 most up-weighted examples in the BiasedSST validation set with a parametric (R-DRO) and non-parametric adversary.}
    \label{tab:my_label}
\end{table}

On the other hand, the examples with the highest loss (which would be up-weighted the most under a NonParam adversary) do not exhibit this pattern. To us, these seem to represent more difficult examples. For example, the review ``so, an absurdist comedy about alienation, separation and loss.'' does not exhibit clear negative sentiment. Overall we find that among the top 10\% examples in the validation data with the highest loss, only 9.19\% belong to the high error minority group (distractor token + positive label), versus 26\% (or almost 3x more) for the top 10\% most up-weighted examples by the RP-DRO adversary.

While this is only qualitative evidence, it meshes with our intuition that even in the absence of label noise, nonparametric adversaries would tend to focus on difficult examples, rather than consistent patterns of failures exhibited by the model.

\section{Effect of Oracle Stopping on Renormalization Strategies}
\label{sec:self_norm_oracle}

\begin{figure}[!h]
\centering
\begin{subfigure}[t]{\columnwidth}
\centering
\includegraphics[width=\columnwidth]{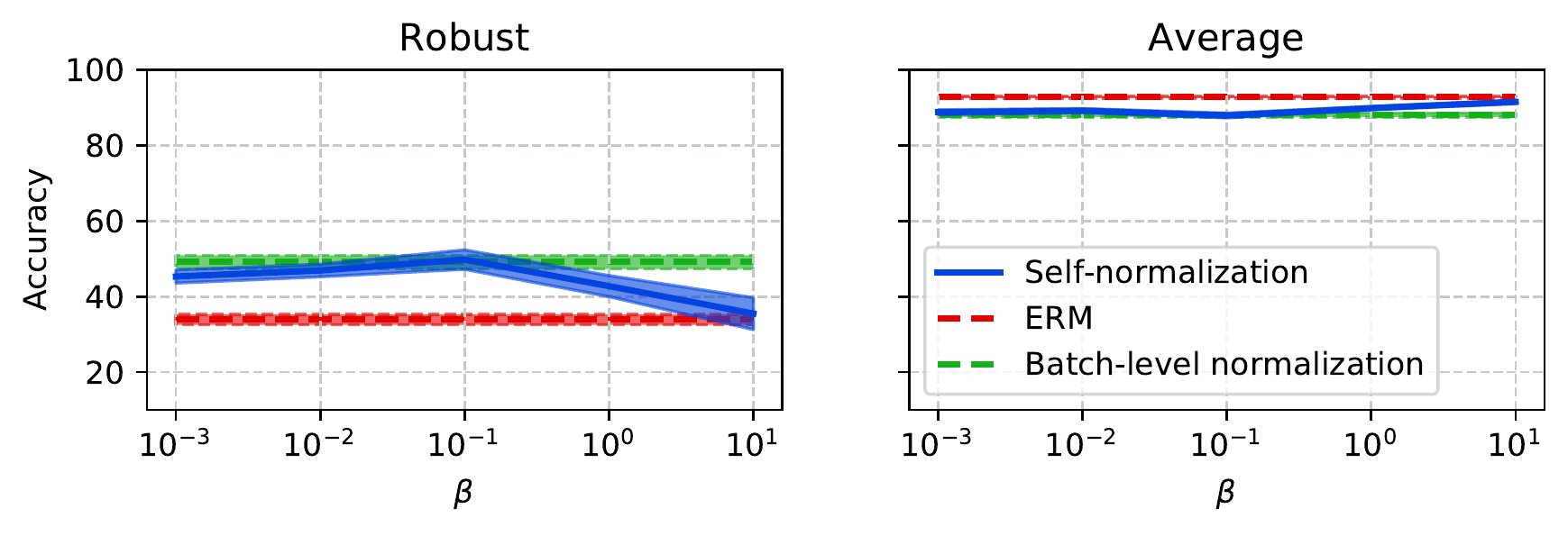}
\end{subfigure}
\caption{\label{fig:self_norm_oracle} Effect of self-normalization coefficient $\beta$ on robust and average accuracy using Oracle stopping. We report results of ERM (which corresponds to $\beta=\infty$) and batch level renormalization for comparison.}
% \vspace{-1em}
\end{figure}

In Figure \ref{fig:self_norm_oracle}, we report the evolution of robust and average accuracy for a model trained with RP-DRO using a self -normalization penalty with a coefficient $\beta$ varying from $10^-3$ to $10^1$.

\section{Stability of Simultaneous Gradient Descent vs. Exact Minmax}
\label{sec:simultaneous_gd_stability}

In the toy setting of Section \ref{sec:simultaneous_gd}, when both the model and the adversary are linear, the resulting min-max problem becomes convex-concave. This means that any stationary point $\theta^*,\psi^*$ of simultaneous gradient descent will be a global saddle-point of the $\mathcal{L}_{\text{RP-DRO}}(\theta,\psi)$ objective. Why then is simultaneous gradient descent systematically achieving higher robust accuracy than the more ``exact case'' where we take gradient steps on the $\max_\psi\Ell_{\text{RP-DRO}}(\theta, \psi)$?

\begin{figure*}[t!]
\centering
\begin{subfigure}[t]{0.3\textwidth}
\centering
\includegraphics[width=\columnwidth]{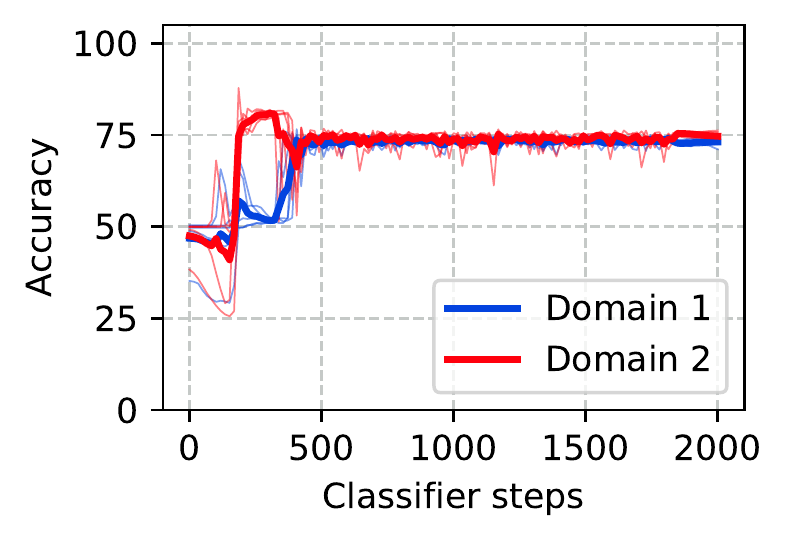}
\caption{\label{fig:linear_1_adv_steps} 1 adversary step (simultaneous updates).
}
\end{subfigure}
~
\begin{subfigure}[t]{0.3\textwidth}
\centering
\includegraphics[width=\columnwidth]{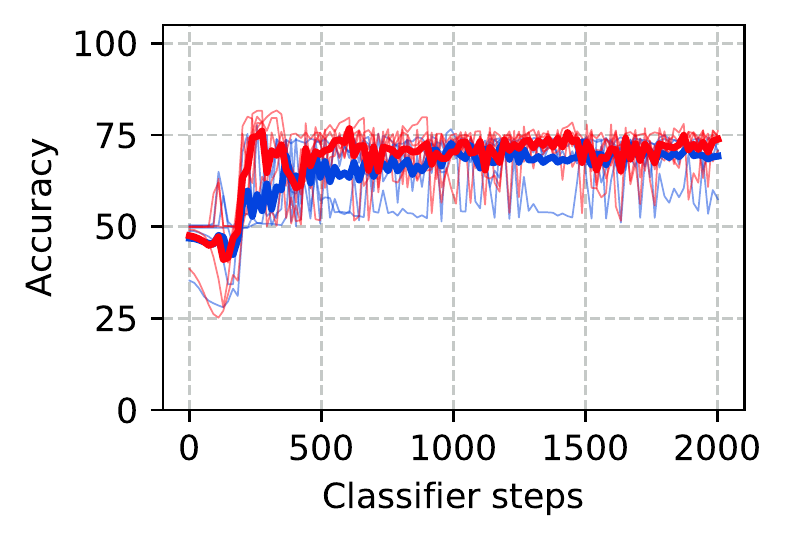}
\caption{\label{fig:linear_10_adv_steps} 10 adversary steps.
}
\end{subfigure}
~
\begin{subfigure}[t]{0.3\textwidth}
\centering
\includegraphics[width=\columnwidth]{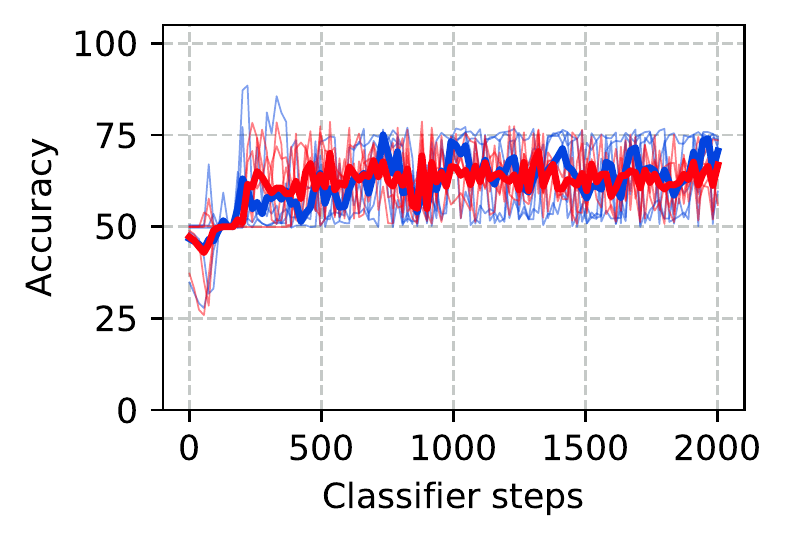}
\caption{\label{fig:linear_100_adv_steps} 100 adversary steps.
}
\end{subfigure}
\caption{\label{fig:adv_steps_traj} Evolution of training trajectories for in our toy setting as the linear adversary is able to take more steps than the classifier. We report accuracies on the two domains for multiple restarts. The bold curves correspond to the average trajectory across all seeds.}
% \vspace{-2em}
\end{figure*}

We find that the benefit of simultaneous updates lies in increasing the stability of training. In Figure \ref{fig:adv_steps_traj}, we report training curves in terms of accuracies on each of the two domains (for 5 random seeds). We vary the number of steps that the adversary is allowed to take in between each classifier update from 1 (simultaneous updates) to 100 (which is closer to ``exact'' case of taking descent steps on $\max_\psi\Ell_{\text{RP-DRO}}(\theta, \psi)$).

We observe that models trained with simultaneous gradient updates consistently converge to the global optimum where both domains achieve the same accuracy. On the other hand the more ``exact'' variants (where the adversary is allowed to take more steps) is much less stable. In the most extreme case (100 adversary steps for each classifier step) the model fails to converge.

\begin{figure*}[t!]
\centering
\begin{subfigure}[t]{0.3\textwidth}
\centering
\includegraphics[width=\columnwidth]{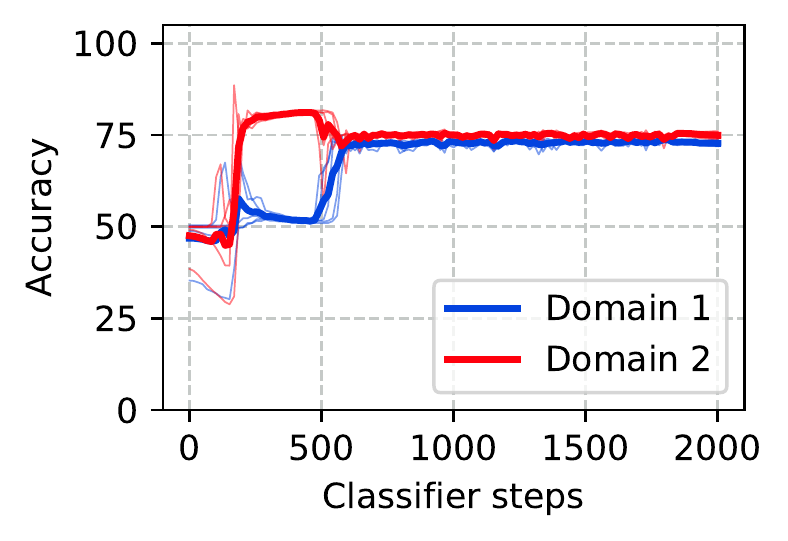}
\caption{\label{fig:mlp_1_adv_steps} 1 adversary step (simultaneous updates).
}
\end{subfigure}
~
\begin{subfigure}[t]{0.3\textwidth}
\centering
\includegraphics[width=\columnwidth]{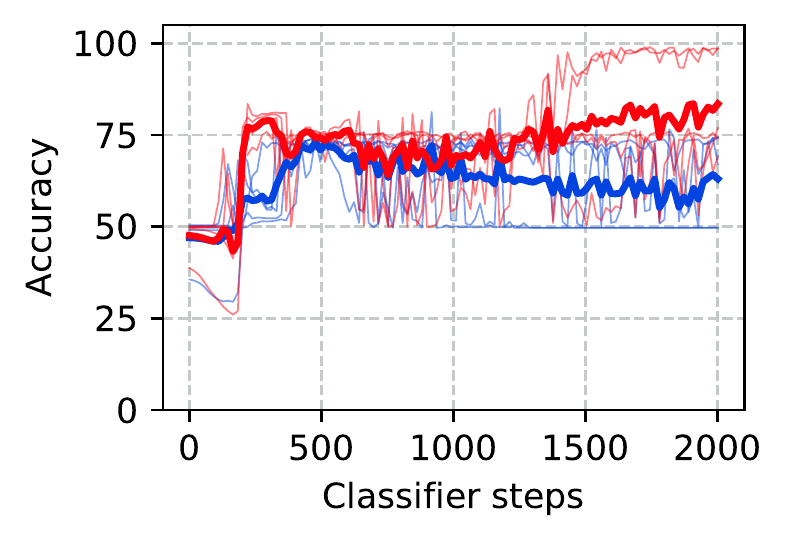}
\caption{\label{fig:mlp_10_adv_steps} 10 adversary steps.
}
\end{subfigure}
~
\begin{subfigure}[t]{0.3\textwidth}
\centering
\includegraphics[width=\columnwidth]{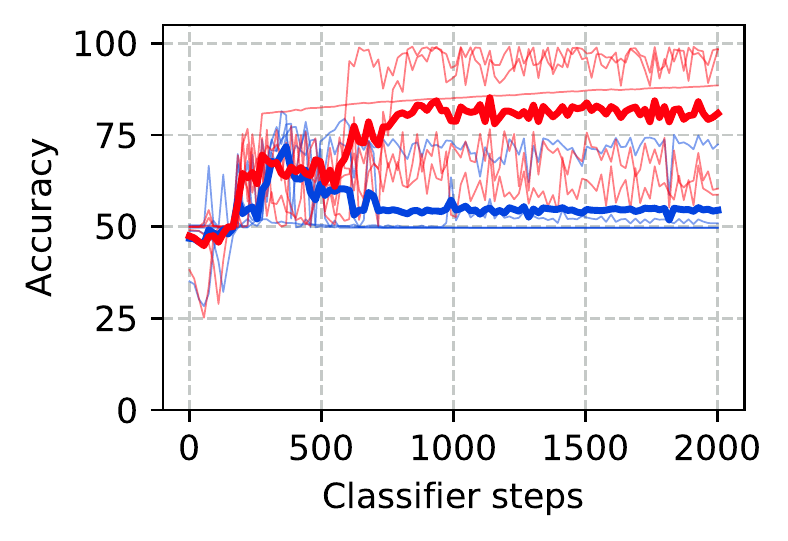}
\caption{\label{fig:mlp_100_adv_steps} 100 adversary steps.
}
\end{subfigure}
\caption{\label{fig:adv_steps_traj_mlp} Evolution of training trajectories for in our toy setting as the MLP-2 adversary is able to take more steps than the classifier. We report accuracies on the two domains for multiple restarts. The bold curves correspond to the average trajectory across all seeds.}
% \vspace{-2em}
\end{figure*}

When models are bigger, and we lose the convex-concavity of the problem, we find that taking steps on $\max_\psi\Ell_{\text{RP-DRO}}(\theta, \psi)$ is also unstable, and sometimes converges to worse local optima compared to simultaneous updates (see Figure \ref{fig:adv_steps_traj_mlp}).

\end{document}